\documentclass{article}

\usepackage{microtype}
\usepackage{graphicx}
\usepackage{subfigure}
\usepackage{booktabs} 

\usepackage[ruled,vlined,linesnumbered]{algorithm2e}

\usepackage{hyperref}

\usepackage[accepted]{icml2025}

\usepackage{amsmath}
\usepackage{amssymb}
\usepackage{mathtools}
\usepackage{amsthm}
\usepackage{amsmath, amssymb}

\usepackage[capitalize,noabbrev]{cleveref}

\theoremstyle{plain}
\newtheorem{theorem}{Theorem}[section]
\newtheorem{proposition}[theorem]{Proposition}
\newtheorem{lemma}[theorem]{Lemma}

\theoremstyle{definition}

\newtheorem{assumption}[theorem]{Assumption}
\theoremstyle{remark}

\usepackage[textsize=tiny]{todonotes}

\usepackage{pifont}

\usepackage[table,xcdraw]{xcolor}
\usepackage{multirow}
\usepackage{adjustbox}
\usepackage{subcaption}
\newcommand{\maj}{\mathrm{maj}}
\newcommand{\hete}{\mathrm{hete}}

\icmltitlerunning{Submission and Formatting Instructions for ICML 2025}

\begin{document}

\twocolumn[
\icmltitle{BoostFGL: Boosting Fairness in Federated Graph Learning}


\begin{icmlauthorlist}
\icmlauthor{Zekai Chen}{yyy}
\icmlauthor{Kairui Yang}{yyy}
\icmlauthor{Xunkai Li}{yyy}
\icmlauthor{Henan Sun}{sch}
\icmlauthor{Zhihan Zhang}{yyy}
\icmlauthor{Jia Li}{sch}
\icmlauthor{Qiangqiang Dai}{yyy}
\icmlauthor{Rong-Hua Li}{yyy}
\icmlauthor{Guoren Wang}{yyy}
\end{icmlauthorlist}

\icmlaffiliation{yyy}{Department of Computer Science, Beijing Institute of Technology, Beijing, China}
\icmlaffiliation{sch}{The Hong Kong University of Science and Technology (GZ), Guangzhou, China}

\icmlcorrespondingauthor{Rong-Hua Li}{lironghuabit@126.com}
\icmlkeywords{Machine Learning, ICML}

\vskip 0.3in
]



\printAffiliationsAndNotice{\icmlEqualContribution} 

\begin{abstract}
Federated graph learning (FGL) enables collaborative training of graph neural networks (GNNs) across decentralized subgraphs without exposing raw data. While existing FGL methods often achieve high overall accuracy, we show that this average performance can conceal severe degradation on disadvantaged node groups. From a fairness perspective, these disparities arise systematically from three coupled sources: label skew toward majority patterns, topology confounding in message propagation, and aggregation dilution of updates from hard clients.
To address this, we propose \textbf{BoostFGL}, a boosting-style framework for fairness-aware FGL. BoostFGL introduces three coordinated mechanisms: \ding{182} \emph{Client-side node boosting}, which reshapes local training signals to emphasize systematically under-served nodes; \ding{183} \emph{Client-side topology boosting}, which reallocates propagation emphasis toward reliable yet underused structures and attenuates misleading neighborhoods; and \ding{184} \emph{Server-side model boosting}, which performs difficulty- and reliability-aware aggregation to preserve informative updates from hard clients while stabilizing the global model. Extensive experiments on 9 datasets show that BoostFGL delivers substantial fairness gains, improving Overall-F1 by 8.43\%, while preserving competitive overall performance against strong FGL baselines.
\end{abstract}

\section{Introduction}
\label{sec:intro}

Graphs naturally model complex interactions among real-world entities and are widely used in domains such as recommendation, finance, and biomedicine~\citep{recommend_system,qiu2023app_gnn_fina3,qu2023app_gnn_bio2}. Graph neural networks (GNNs) exploit these relational structures to learn expressive node representations for downstream tasks~\citep{kipf2017semi,velickovic2018graph}. In many practical settings, however, graph data are distributed across organizations and cannot be centralized due to privacy and regulatory constraints. Federated graph learning (FGL) enables collaborative training over decentralized subgraphs and has emerged as a promising paradigm to balance utility and privacy~\citep{FGLsurvey,wu2025comprehensive}.

Most existing FGL methods are primarily evaluated by overall accuracy and can appear successful under this metric. Yet, strong average performance can coexist with substantial degradation on under-represented node groups, such as minority classes, heterophilous nodes, and their intersection (Table~\ref{table:node_cls}). While similar disparities have been reported in centralized graph learning, federation can further amplify them: decentralized optimization and cross-client aggregation systematically concentrate learning capacity on statistically and structurally ``easy'' regions, weakening corrective signals for disadvantaged nodes. We substantiate this amplification with process-level diagnostics in Sec.~\ref{sec:empirical}, which reveal skewed gradient allocation (Fig.~\ref{fig:gsd_curve}), unreliable message propagation in heterophilous/minority regions (Fig.~\ref{fig:epr_dist}), and update cancellation under uniform aggregation (Fig.~\ref{fig:dr_ratio_overview}). A fairness-centered examination of FGL is therefore needed---not only \emph{whether} the global model is accurate on average, but \emph{for whom} this accuracy is achieved.

Despite strong average accuracy, fairness remains problematic because improvements are often unevenly distributed across both populations and clients. From a \emph{where-it-happens} perspective, the causes arise at different stages of the client--server pipeline. On the \textbf{client side}, \emph{label skew at the local optimization level} emerges when imbalanced labels bias local objectives toward majority patterns, so minority and hard cases receive weak corrective signals; recent fairness-aware FGL designs partially address this by re-balancing node supervision or objectives~\cite{fairfgl}. Also on the client side, \emph{topology confounding at the message passing level} further amplifies disparity: in heterophilous or structurally sparse regions, neighborhood aggregation mixes inconsistent signals and disadvantaged nodes receive less reliable propagation, motivating topology-aware corrections in both fairness-centered FGL and heterophily-oriented GNN studies~\cite{fairfgl,Zheng2022HeterophilySurvey,Zhu2021GNNHeterophily}. On the \textbf{server side}, \emph{aggregation dilution at the global aggregation level} arises because clients rich in minority or heterophilous structures tend to yield higher-variance, harder-to-optimize updates that can be washed out by easier clients under uniform averaging, a concern related to fairness-aware aggregation principles studied in federated learning beyond graphs~\cite{Mohri2019AgnosticFL,Li2019qFFL,Ezzeldin2023FairFed,Wang2021FedFV}. Taken together, existing approaches typically intervene at only one part of this pipeline (local reweighting, topology correction, or server aggregation), leaving a gap for a coordinated and general solution that is compatible with standard FGL workflows and can \emph{complement} specialized fairness architectures.

To address these coupled issues, we propose \textbf{BoostFGL}, a \emph{boosting-inspired} and modular framework for fairness-aware FGL that can be plugged into mainstream client--server training. BoostFGL introduces three coordinated mechanisms aligned with the above loci: \textbf{\ding{182} Client-side node boosting} enhances the learning signal for systematically under-served nodes without changing label support; \textbf{\ding{183} Client-side topology boosting} reallocates propagation emphasis toward reliable yet underused structures while attenuating misleading neighborhoods; and \textbf{\ding{184} Server-side model boosting} performs difficulty- and reliability-aware aggregation to preserve informative updates from hard clients and stabilize global training. Importantly, BoostFGL is orthogonal to backbone architectures and optimization choices, and can be stacked on top of existing FGL methods (e.g., personalized optimization or fairness-centered designs) to further reduce group disparities.

\textbf{Our contributions} are threefold.
\underline{\textbf{\textit{(1) New Perspective.}}} We provide a unified analysis of unfairness in FGL by decomposing biased training dynamics along the client--server pipeline into label skew, topology confounding, and aggregation dilution, and relate them to concrete disadvantaged node groups (e.g., minority and heterophilous nodes).
\underline{\textbf{\textit{(2) New Framework.}}} We propose BoostFGL, a modular and composable framework that jointly corrects supervision, message passing, and aggregation biases, while remaining compatible with mainstream FGL protocols and architectures.
\underline{\textbf{\textit{(3) SOTA Performance.}}} Extensive experiments on 9 datasets show that BoostFGL improves Overall-F1 by \textbf{8.43\%} and Accuracy by \textbf{2.46\%} over strong baselines. It also remains robust on large-scale graphs where several competitive methods encounter memory limits (OOM).

\section{Preliminaries}
\label{sec:prelim}

\subsection{Problem Formulation}

We study semi-supervised node classification under the subgraph-federated learning (subgraph-FL) setting.
Let $\mathcal{G}=(\mathcal{V},\mathcal{E},X,Y)$ be an attributed graph with $X\in\mathbb{R}^{|\mathcal{V}|\times d}$ and $Y\in\{1,\ldots,C\}^{|\mathcal{V}|}$.
Denote labeled nodes by $\mathcal{V}^L$ and unlabeled nodes by $\mathcal{V}^U$.
A message-passing GNN $f_\theta$ predicts class probabilities for node $v$ using its local neighborhood.
In subgraph-FL, $\mathcal{G}$ is partitioned over $K$ clients, where client $k$ holds a local subgraph
$\mathcal{G}_k=(\mathcal{V}_k,\mathcal{E}_k,X_k,Y_k)$ with $\bigcup_{k=1}^K \mathcal{V}_k=\mathcal{V}$ and $\mathcal{V}_i\cap\mathcal{V}_j=\emptyset$ for $i\neq j$.
Training and message passing are performed locally, and only model updates are communicated.
Let $\mathcal{V}_k^L=\mathcal{V}_k\cap\mathcal{V}^L$ and $n^L=|\mathcal{V}^L|$.
The standard objective is the weighted empirical risk
\begin{equation}
\label{eq:boostfgl-fedobj}
\begin{aligned}
\min_{\theta}\ \mathcal{L}_{\mathrm{fed}}(\theta)
&:= \frac{1}{n^L}\sum_{k=1}^K\sum_{v\in\mathcal{V}_k^L}
\ell\!\left(f_\theta(v;\mathcal{G}_k),y_v\right).
\end{aligned}
\end{equation}

where $\ell$ is cross-entropy and $\mathcal{L}_k$ is the local empirical risk.
As we empirically show in Sec.~\ref{sec:empirical}, optimizing Eq.~\eqref{eq:boostfgl-fedobj} can systematically under-allocate training signal to minority and heterophilous nodes, yielding group-wise performance gaps.

\vspace{-8pt}

\subsection{Related Work}

\paragraph{Graph neural networks.}
GNNs predict node labels via message passing, i.e., iterative neighborhood propagation and aggregation~\citep{kipf2017semi,hamilton2017inductive,velickovic2018graph}. 
Under heterophily and label imbalance, vanilla propagation may be unreliable or even harmful as neighbors inject misleading signals~\citep{pei2020geomgcn}. 
Rather than redesigning propagation rules for centralized training, we treat GNNs as local encoders and analyze how federated optimization and aggregation exacerbate or correct these biases across decentralized subgraphs.

\vspace{-8pt}

\paragraph{Federated graph learning.}
FGL extends federated learning to graph-structured data, typically under the subgraph-FL paradigm where each client owns a local subgraph.
Existing FGL methods primarily address utility degradation caused by non-IID data and structural incompleteness.
Representative approaches include neighbor reconstruction or completion~\citep{zhang2021subgraph}, personalized heads or clustering-based aggregation~\citep{baek2023personalized,li2024fedgta}, and graph condensation or prototype sharing to improve efficiency and generalization~\citep{chen2025fedc4}.
These methods modify local objectives, message passing, or aggregation rules to improve \emph{average accuracy}.
However, they typically optimize global performance metrics and do not explicitly reason about node-level disparity induced by label imbalance neighborhoods.
As a result, improvements in overall accuracy may conceal systematic degradation on minority or structurally disadvantaged nodes.

\paragraph{Fairness in federated learning.}
Fairness has been extensively studied in Euclidean federated learning, predominantly at the client level.
Agnostic federated learning optimizes worst-case mixtures of client distributions~\citep{mohri2019agnostic}, while q-FFL and related methods reweight client updates to reduce disparity across devices~\citep{li2020fair,Wang2021FedFV}.
Other approaches modify aggregation or sampling strategies to favor under-served clients~\citep{Ezzeldin2023FairFed}.
These methods assume i.i.d.\ samples within each client and do not consider graph-specific phenomena such as message-passing bias, neighborhood-induced label leakage, or topology-driven error amplification.
Consequently, their fairness notions do not directly transfer to node-level fairness in graph-structured federated settings.

\vspace{-15pt}

\paragraph{Fairness-aware federated graph learning.}
Fairness-aware FGL has only recently begun to receive attention.
Some robust or cross-domain FGL approaches share prototypes or structural proxies to mitigate distribution shift~\citep{wan2024fgpproto,fu2024fedspray}, where fairness often improves implicitly as a side effect of robustness.
FairFGL~\citep{fairfgl} takes an important step by explicitly operationalizing fairness from both label- and topology-aware perspectives and proposing targeted interventions for minority and heterophilous nodes.
Nevertheless, a broader challenge remains: in subgraph-FL, unfairness is shaped by \emph{coupled} dynamics across the client--server pipeline, where optimization, message passing, and aggregation interact over rounds.
As a result, addressing a single stage in isolation may yield incomplete or unstable gains when the other stages continue to amplify disparity.
This motivates \emph{pipeline-level} treatments that coordinate corrections across local objectives, propagation, and server aggregation.

\vspace{-6pt}

\subsection{Positioning of BoostFGL.}
BoostFGL complements existing fairness-aware FGL methods by providing a \emph{plug-in} framework that targets unfairness \emph{where it emerges} along the client--server pipeline.
Unlike approaches that redesign a specific local model or aggregation rule, our boosting modules are lightweight, orthogonal to backbone choices, and can be integrated into standard subgraph-FL training to jointly mitigate coupled biases from optimization, propagation, and aggregation.

\begin{figure}[t]
  \centering
  \includegraphics[width=\linewidth]{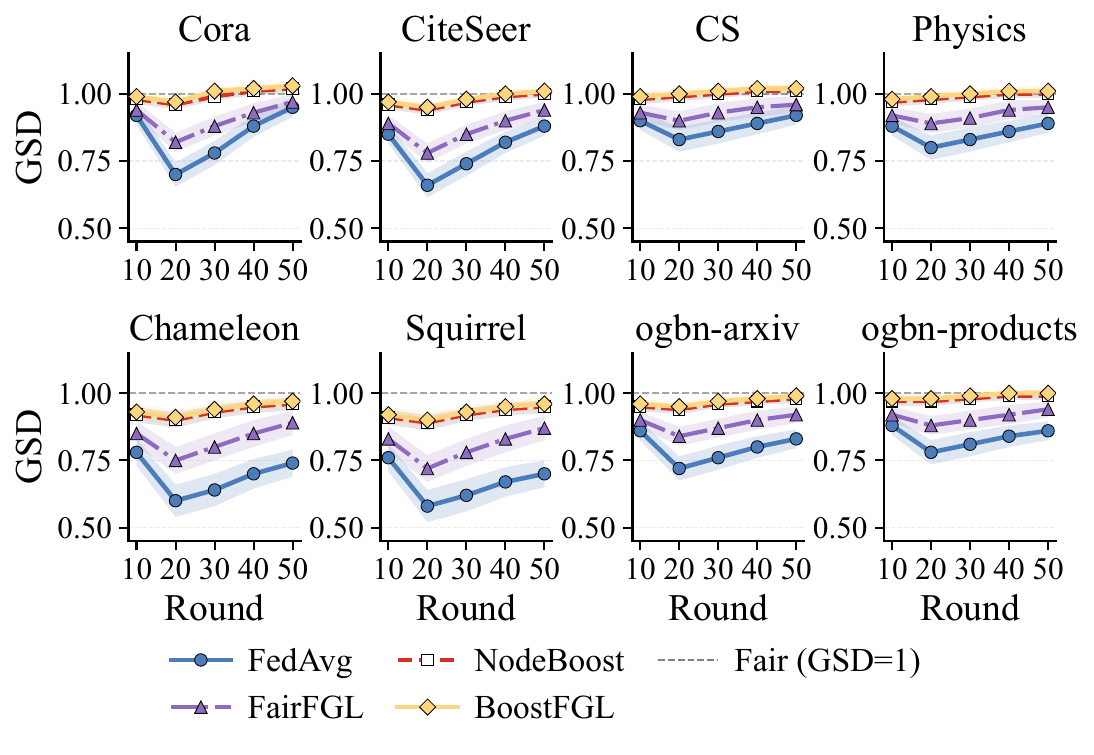}
  \caption{\textbf{Label skew diagnosis.} Gradient Share Disparity (GSD) over federated rounds. The dashed line denotes the fair allocation baseline (GSD$=1$).}
  \label{fig:gsd_curve}
\end{figure}

\begin{figure*}[t]
  \centering
  \begin{minipage}[t]{0.48\linewidth}
    \centering
    \includegraphics[width=\linewidth]{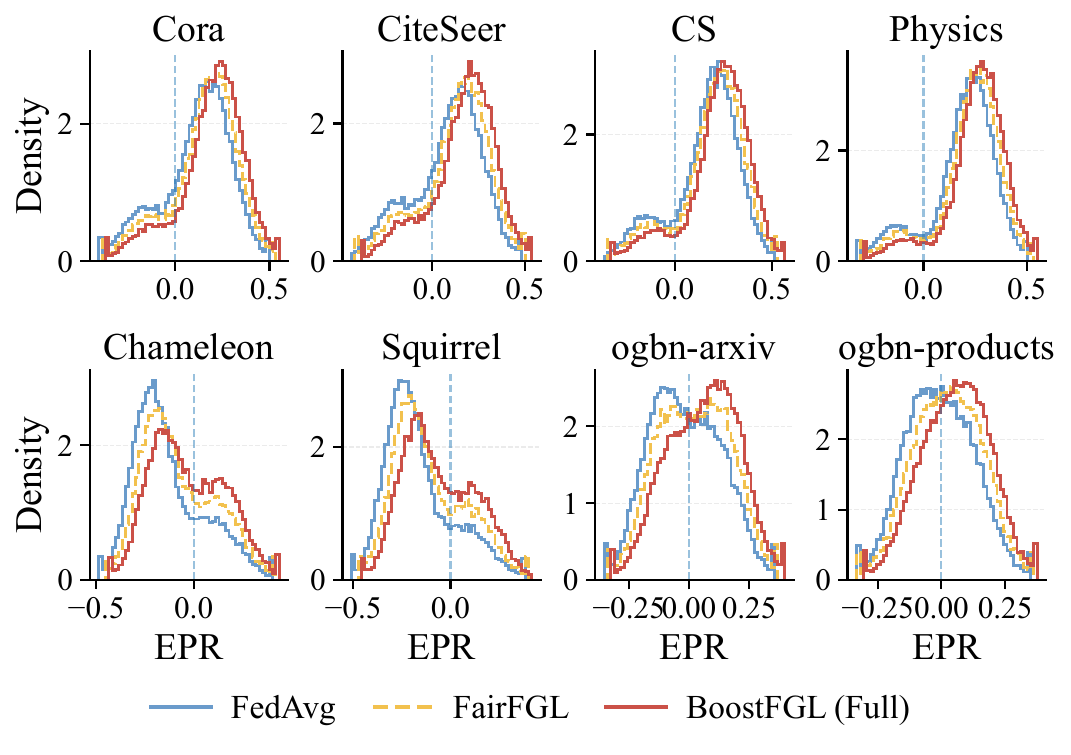}
    \caption{\textbf{Topology confounding diagnosis.}
    Distribution of Edge-wise Propagation Reliability (EPR).
    Negative EPR indicates harmful message passing.}
    \label{fig:epr_dist}
  \end{minipage}
  \hfill
  \begin{minipage}[t]{0.48\linewidth}
    \centering
    \includegraphics[width=\linewidth]{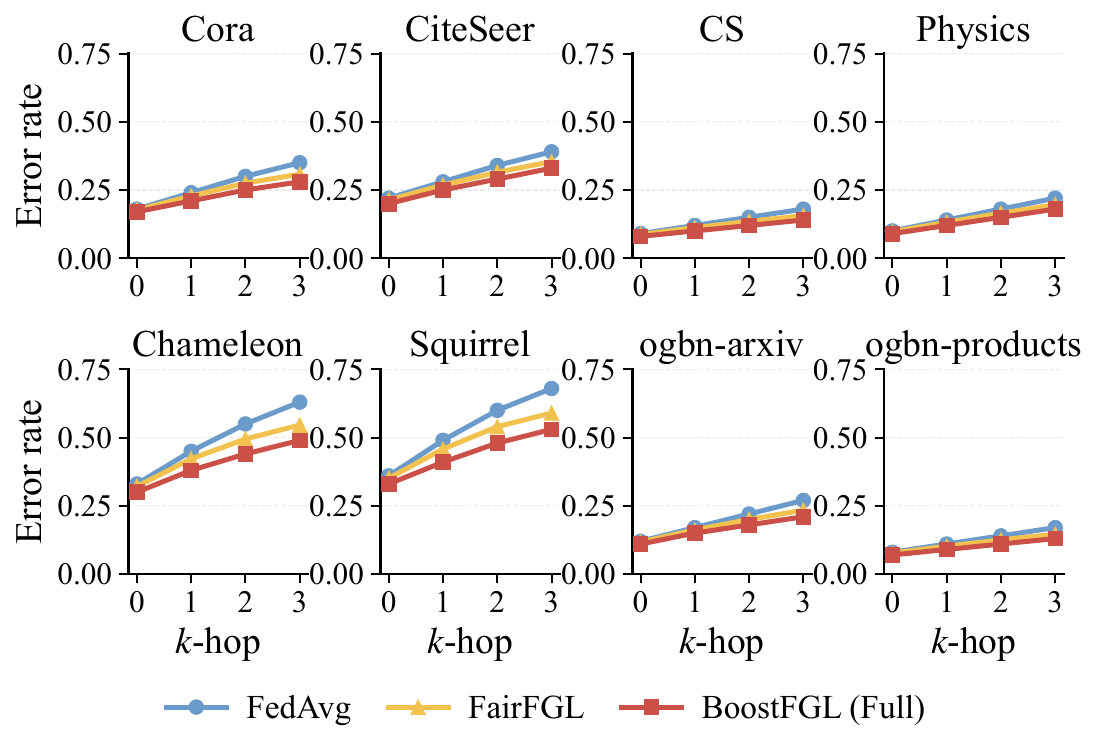}
    \caption{\textbf{K-hop error amplification.}
    Error propagation as a function of distance from minority/heterophilous nodes.
    BoostFGL suppresses amplification under message passing.}
    \label{fig:khop_amp}
  \end{minipage}
\end{figure*}

\vspace{-6pt}

\section{Empirical Study \& Theoretical Analysis}
\label{sec:empirical}

We verify that the three unfairness sources in Sec.~\ref{sec:intro}---\emph{label skew}, \emph{topology confounding}, and \emph{aggregation dilution}---appear systematically in subgraph-FL.
Instead of only reporting end-task metrics, we introduce \emph{process-level diagnostics} to probe optimization, propagation, and aggregation dynamics.
All diagnostics are computed on the same federated subgraphs used for training and reported in aggregated form; more details are applied in Appendix~\ref{app:additional_empirical}.

\vspace{-8pt}

\subsection{Node-side label skew}
\label{sec:empirical_node}
Fig.~\ref{fig:gsd_curve} reports Gradient Share Disparity (GSD) over rounds.
During training, GSD stays below $1$, indicating systematic under-allocation of gradient mass to minority/hard nodes.
This behavior is consistent across minority definitions and client heterogeneity, motivating node-side boosting.

\subsection{Topology Confounding in Message Passing}
\label{sec:empirical_topology}
Fig.~\ref{fig:epr_dist} shows the distribution of Edge-wise Propagation Reliability (EPR), where negative values indicate harmful messages.
Edges around minority/heterophilous regions exhibit a heavier negative tail, and Fig.~\ref{fig:khop_amp} further shows error amplification with increasing hop distance.
These results highlight a structural propagation issue that motivates topology-side boosting beyond objective reweighting.

\subsection{Server-side Aggregation Dilution}
\label{sec:empirical_agg}
We evaluate whether uniform aggregation cancels fairness-improving updates.
The Dilution Ratio (DR) in Fig.~\ref{fig:dr_ratio_overview} is low under standard averaging, consistent with destructive interference among heterogeneous client updates.
This motivates trust-aware aggregation in model-side boosting.

\subsection{Theoretical Analysis}
\label{sec:theory}

We provide diagnostic-aligned guarantees showing that the three boosting modules in \textbf{BoostFGL}
monotonically improve the same process-level signals measured in Sec.~\ref{sec:empirical}:
GSD, EPR, DR.

\paragraph{Node-side boosting increases gradient-share parity (GSD).}
Recall the node weight $\alpha_v^{(t)} \triangleq 1 + \lambda_n\,\bar d_v^{(t)}$ (Eq.~\eqref{eq:node_weight}).
Let $g_v^{(t)} := \nabla_\theta \ell(v;\theta)$ be the per-node gradient and define the per-group gradient mass
$G_g^{(t)} := \sum_{v \in \mathcal{V}_g^L} \|g_v^{(t)}\|_2$ on labeled nodes, where $g \in \{\text{min}, \text{maj}\}$ denotes minority vs.\ majority.
We use the diagnosis
\[
\mathrm{GSD}^{(t)} := 
\frac{G_{\mathrm{min}}^{(t)}/\lvert \mathcal{V}_{\mathrm{min}}^L \rvert}
     {G_{\mathrm{maj}}^{(t)}/\lvert \mathcal{V}_{\mathrm{maj}}^L \rvert}.
\]

\begin{lemma}[Gradient-share rectification]
\label{lem:gsd_rect}
Let $\mathrm{GSD}_{\textsc{base}}^{(t)}$ be computed under uniform weighting ($\alpha_v^{(t)} \equiv 1$), and
$\mathrm{GSD}_{\textsc{boost}}^{(t)}$ under BoostFGL node weights.
Under mild difficulty--gradient coupling and $\mathbb{E}[\bar d \mid \min] > \mathbb{E}[\bar d \mid \maj]$,

\vspace{-5pt}

\begin{equation}
\label{eq:gsd_rect}
\mathrm{GSD}_{\textsc{boost}}^{(t)}
\ \ge\
\mathrm{GSD}_{\textsc{base}}^{(t)}
\cdot
\frac{1+\lambda_n \,\mathbb{E}[\bar d \mid \min]}{1+\lambda_n \,\mathbb{E}[\bar d \mid \maj]}.
\end{equation}
\end{lemma}

\vspace{-8pt}

\noindent\emph{Use.} Lemma~\ref{lem:gsd_rect} formalizes that increasing $\lambda_n$ amplifies minority/hard-node gradients
relative to majority, pushing GSD toward the fair baseline (Fig.~\ref{fig:gsd_curve}).

\vspace{-8pt}

\paragraph{Topology-side boosting suppresses harmful messages (negative EPR).}
Recall $\beta_{uv}^{(t)} \propto \exp(\lambda_e s_{uv}^{(t)})$ (Eq.~\eqref{eq:edge_weight}).
For a target node $v$, view neighbor sampling as drawing $u \sim q_{\lambda_e}(\cdot\mid v)$ with
$q_{\lambda_e}(u\mid v) := \beta_{uv}^{(t)}$.
Let $R_{uv}^{(t)}$ denote the (signed) EPR of message $(u\!\to\! v)$ at round $t$.

\begin{theorem}[Harmful-message suppression]
\label{thm:epr_suppress}
Assume the score $s_{uv}^{(t)}$ is an affine noisy proxy of reliability, i.e.,
$s_{uv}^{(t)}=\kappa R_{uv}^{(t)}+\xi_{uv}$ with $\kappa>0$ and $\xi_{uv}$ sub-Gaussian.
Then for any fixed $v$ and round $t$,
\begin{equation}
\label{eq:epr_mean_mono}
\mathbb{E}_{u\sim q_{\lambda_e}(\cdot\mid v)}\!\left[R_{uv}^{(t)}\right]
\ \text{is non-decreasing in }\lambda_e.
\end{equation}
and the harmful-message probability admits the tail bound

\vspace{-15pt}

\begin{equation} \label{eq:epr_neg_tail} \mathbb{P}_{u\sim q_{\lambda_e}(\cdot\mid v)}\!\left(R_{uv}^{(t)}<0\right)\le\exp\!\left(-\lambda_e \kappa\, \bar R_v^{(t)} + \tfrac{\lambda_e^2\sigma^2}{2}\right). \end{equation}

\vspace{-13pt}

where $\bar R_v^{(t)} := \mathbb{E}_{u\sim q_{\lambda_e}(\cdot\mid v)}[R_{uv}^{(t)}]$ and $\sigma^2$ is the
sub-Gaussian proxy variance.
\end{theorem}

\vspace{-11pt}

\noindent\emph{Use.} Theorem~\ref{thm:epr_suppress} explains why exponential tilting reduces the negative-EPR tail and improves
average propagation reliability (Fig.~\ref{fig:epr_dist}).

\vspace{-8pt}

\paragraph{Model-side boosting mitigates aggregation dilution (DR).}
Let $\mathcal{M}_t$ be participating clients and define the minority-descent direction
$\mathbf{g}_{\min}^{(t)}$ (estimated as in Fig.~\ref{fig:dr_ratio_overview}).
Let $a_m^{(t)} := \langle \Delta\theta_m^{(t)}, \mathbf{g}_{\min}^{(t)} \rangle$ be the alignment.
We measure dilution by comparing the aggregated update's projection onto $\mathbf{g}_{\min}^{(t)}$
against the average positive alignment across clients, yielding a scale-normalized ratio:

\vspace{-10pt}

\begin{equation}
\label{eq:dr_def_main}
\mathrm{DR}^{(t)}
:=
\frac{
\left\langle
\sum_{m\in\mathcal{M}_t} w_m^{(t)}\Delta\theta_m^{(t)},\
\mathbf{g}_{\min}^{(t)}
\right\rangle
}{
\frac{1}{|\mathcal{M}_t|}\sum_{m\in\mathcal{M}_t}\left[a_m^{(t)}\right]_+ + \epsilon
}.
\end{equation}

\vspace{-8pt}

\noindent
where $[a_m^{(t)}]_+ \triangleq \max\{a_m^{(t)},0\}$ denotes the positive part,
$w_m^{(t)}$ are aggregation weights, and $\epsilon>0$ is a small constant.

\begin{theorem}[Trust-gated aggregation improves DR and bounds influence]
\label{thm:dr_improve}
Let $w_m^{(t)} \propto \tau_m^{(t)}$ with $\tau_m^{(t)}$ from Eq.~\eqref{eq:trust}.
Assume $\tau_m^{(t)}$ is non-decreasing in $[a_m^{(t)}]_+$ in expectation.
Then, for each round $t$,

\vspace{-14pt}

\begin{equation}
\label{eq:dr_improve_main}
\mathrm{DR}_{\textsc{boost}}^{(t)} \ \ge\ \mathrm{DR}_{\textsc{fedavg}}^{(t)}.
\end{equation}
Moreover, each client’s effective influence is uniformly bounded as
\begin{equation}
\label{eq:influence_bound_main}
\big\|w_m^{(t)}\Delta\theta_m^{(t)}\big\|_2
\ \le\
\frac{1}{\lambda_s \sum_{j\in\mathcal{M}_t}\tau_j^{(t)}}.
\end{equation}

\vspace{-10pt}

\end{theorem}
\noindent\emph{Use.} Theorem~\ref{thm:dr_improve} formalizes that trust weights preserve minority-improving signal
(and avoid destructive interference), matching the DR gains in Fig.~\ref{fig:dr_ratio_overview}.


\vspace{-10pt}

\paragraph{Asymptotic consistency.}
In the high-confidence stationary regime, the boosting factors vanish and the procedure collapses to uniform aggregation:

\begin{proposition}[Reduction to standard FedAvg]
\label{prop:consistency}
If training reaches a high-confidence stationary regime where $\bar d_v^{(t)}\!\to\!0$ and update magnitudes vanish,
then $\alpha_v^{(t)}\!\to\!1$, $\tau_m^{(t)}\!\to\!1$, and BoostFGL reduces to standard FedAvg asymptotically.
All assumptions and detailed proofs are deferred to Appendix~\ref{app:proofs}.
\end{proposition}

\begin{figure*}[!t]
  \centering
  \includegraphics[width=\textwidth]{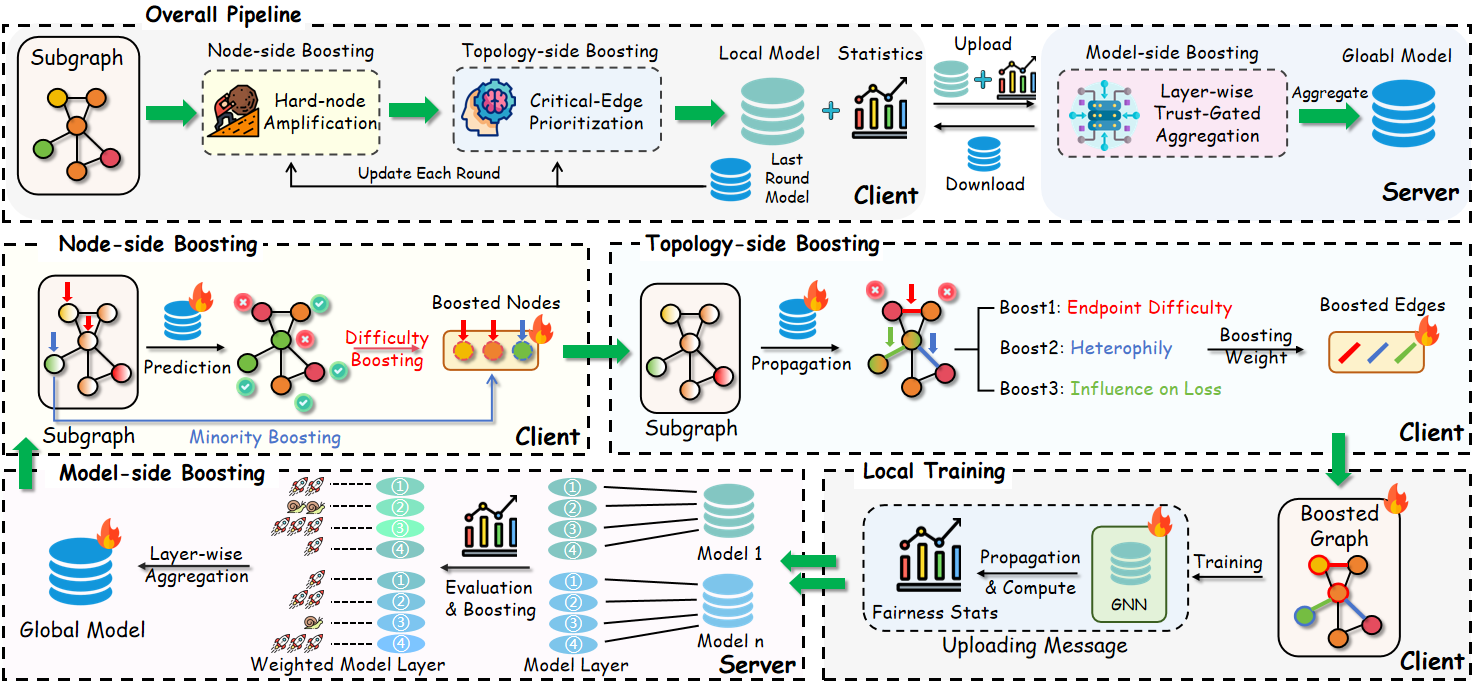}
  \caption{Overview of the proposed \textbf{BoostFGL} framework. 
Each client applies node-side and topology-side boosting to amplify hard nodes and critical edges before local training, then uploads the updated model and statistics. The server performs model-side, layer-wise boosting during aggregation to obtain the global model for the next communication round.}
  \label{fig:framework}
\end{figure*}

\vspace{-15pt}

\section{Methodology}

\subsection{Overview}
We present \textbf{BoostFGL}, a boosting-style framework for fairness-aware FGL (Fig.~\ref{fig:framework}).
BoostFGL augments the standard subgraph-FL pipeline with three lightweight and modular boosting operations:
\emph{Client-side node boosting} reweights local supervision to emphasize systematically under-served nodes,
\emph{Client-side topology boosting} reshapes message propagation by prioritizing fairness-critical edges during message passing,
and \emph{Server-side model boosting} performs trust-gated aggregation on the server to preserve informative updates from hard clients.
All components are executed locally or depend only on aggregated statistics, ensuring full compatibility with standard FGL protocols. The complete client--server procedure is summarized in Appendix~\ref{app:pseudocode}.

\subsection{Client-side Node Boosting}
\label{sec:node_boost}

\textbf{Objective.}
Client-side node boosting mitigates label skew by assigning more optimization weight to systematically under-served nodes
(e.g., minority-class nodes and persistently hard examples) within each client.

\textbf{Node difficulty.}
For a labeled node $v$ with ground-truth label $y_v$, let $p_{\theta_m}(y_v \mid v)$ denote the predicted probability under the local model $\theta_m$.
We define a simple and stable difficulty score and maintain its exponential moving average (EMA) to reduce noise across local updates:

\vspace{-10pt}

\begin{equation}
\label{eq:node_difficulty}
\bar d_v^{(t)}
\coloneqq
\bar d_v^{(t-1)}+\mu\Bigl(\underbrace{1 - p_{\theta_m^{(t)}}(y_v \mid v)}_{\triangleq\, d_v^{(t)}}
-\bar d_v^{(t-1)}\Bigr).
\end{equation}

\vspace{-8pt}

\noindent
We set $\bar d_v^{(0)}=0$ and choose $\mu\in(0,1]$. Since $p_{\theta}(y_v\mid v)\in[0,1]$, we have $\bar d_v^{(t)}\in[0,1]$,
keeping subsequent boosting weights bounded.
For unlabeled nodes (used by topology boosting in Sec.~\ref{sec:topo_boost}), we use a confidence proxy
$d_v^{(t)}\coloneqq 1-\|p_{\theta_m^{(t)}}(\cdot\mid v)\|_{\infty}$ to obtain a compatible $\bar d_v^{(t)}$.

\textbf{Boosted node objective.}
We convert difficulty into a bounded boosting weight

\vspace{-15pt}

\begin{equation}
\label{eq:node_weight}
\alpha_v^{(t)} \triangleq 1 + \lambda_n\,\bar d_v^{(t)}, \qquad \lambda_n \ge 0,
\end{equation}

\vspace{-8pt}

and optimize the reweighted empirical risk on client $m$:
\begin{equation}
\label{eq:node_loss}
\mathcal{L}_m^{\textsc{node}}(\theta)
\coloneqq
\sum_{v \in \mathcal{V}_m^{\mathrm{lab}}}
\alpha_v^{(t)} \,
\ell\!\left(f_\theta(v;\mathcal{G}_m),y_v\right)
\;\equiv\;
\big\langle \alpha^{(t)}, \ell(\theta) \big\rangle_{\mathcal{V}_m^{\mathrm{lab}}}.
\end{equation}

\vspace{-7pt}

where $\ell(\cdot)$ is the standard node classification loss and
$\langle \alpha^{(t)}, \ell(\theta)\rangle_{\mathcal{V}_m^{\mathrm{lab}}}
\triangleq \sum_{v\in\mathcal{V}_m^{\mathrm{lab}}}\alpha_v^{(t)}\,\ell(f_\theta(v;\mathcal{G}_m),y_v)$.
This linear reweighting highlights hard and minority nodes without the instability of exponential weights.

\subsection{Client-side Topology Boosting}
\label{sec:topo_boost}

\textbf{Objective.}
Even with balanced supervision, unfairness may persist due to \emph{topology confounding} during message passing, where uniformly aggregating neighbors injects misleading signals in heterophilous or irregular regions.
Client-side topology boosting addresses this issue by prioritizing propagation along fairness-critical edges.

\textbf{Edge criticality.}
For each (directed) incoming edge $(u\!\to\! v)\in\mathcal{E}_m$, we define an interpretable topology score

\vspace{-22pt}

\begin{equation}
\label{eq:edge_score}
s_{uv}^{(t)} \triangleq \tfrac12\!\big(\bar d_u^{(t)}+\bar d_v^{(t)}\big)+
\begin{cases}
\mathbb{I}[y_u\neq y_v], & u,v\in\mathcal{V}_m^{\mathrm{lab}},\\
1-\langle p_u,p_v\rangle, & \text{otherwise},
\end{cases}
\end{equation}

\vspace{-5pt}

\vspace{-7pt}

which combines endpoint difficulty (from Sec.~\ref{sec:node_boost}) and heterophily.
Here $\mathbb{I}[\cdot]$ is the indicator function, and $p_u,p_v$ are predicted label distributions under the current local model.
This fallback avoids reliance on unavailable labels and preserves applicability in semi-supervised settings.

\textbf{Boosted message passing.}
We normalize topology scores into attention-like propagation weights over the incoming neighborhood of each target node:

\vspace{-12pt}
\begin{equation}
\label{eq:edge_weight}
\beta_{\cdot v}^{(t)} \triangleq \nabla_{\mathbf{s}_{\cdot v}^{(t)}}\,
\frac{1}{\lambda_e}\log\!\sum_{u\in\mathcal{N}(v)}\exp\!\big(\lambda_e s_{uv}^{(t)}\big).
\end{equation}

\vspace{-8pt}

\noindent where $\lambda_e\ge 0$ and $\mathbf{s}_{\cdot v}^{(t)}\!\coloneqq (s_{uv}^{(t)})_{u\in\mathcal{N}(v)}$.
We then incorporate $\beta_{uv}^{(t)}$ into message passing:
\begin{equation}
\label{eq:boosted_mp}
\mathbf{h}_v^{(\ell+1)}
=
\sigma\!\left(
(\beta_{\cdot v}^{(t)})^\top \mathbf{S}_v\,\mathbf{H}^{(\ell)}\,(\mathbf{W}^{(\ell)})^\top
\right),
\end{equation}

\vspace{-10pt}

By prioritizing edges incident to hard or heterophilous nodes, topology-side boosting reallocates propagation capacity toward underrepresented structural patterns and attenuates misleading neighborhoods.





\subsection{Server-side Model Boosting}
\label{sec:model_boost}

\textbf{Objective.}
Aggregation dilution arises when informative but high-variance updates from hard clients are suppressed by uniform averaging.
Server-side model boosting mitigates this effect by assigning trust weights to client updates based on stability and fairness statistics.

Let $\Delta\theta_m^{(t)} \coloneqq \theta_m^{(t)} - \theta^{(t-1)}$ denote the update from client $m$.
Each client additionally uploads a lightweight fairness summary $\mathbf{s}_m^{(t)}$, including the mean difficulty on minority nodes and the local gap between minority and majority accuracy.
We define a bounded trust score
\begin{equation}
\label{eq:trust}
\tau_m^{(t)} \triangleq
\exp\!\Big(
-\lambda_s\|\Delta\theta_m^{(t)}\|_2
-\gamma\,\mathrm{Gap}(\mathbf{s}_m^{(t)})
\Big),
\end{equation}
where $\mathrm{Gap}(\cdot)$ measures the disparity between minority and majority node performance on the client, and $\lambda_s,\gamma\ge0$ control robustness and fairness sensitivity.

\textbf{Trust-gated aggregation.}
The global model is updated via normalized trust-weighted averaging:

\vspace{-25pt}

\begin{equation}
\label{eq:agg}
\theta^{(t)}=\theta^{(t-1)}+\big\langle \boldsymbol{\pi}^{(t)},\,\Delta\boldsymbol{\Theta}^{(t)}\big\rangle,
\boldsymbol{\pi}^{(t)}\coloneqq \frac{\boldsymbol{\tau}^{(t)}}{\mathbf{1}^\top \boldsymbol{\tau}^{(t)}},
\end{equation}

\vspace{-10pt}

This aggregation preserves informative updates from hard clients while maintaining stability across rounds.

\begin{table*}[t]
    \centering
   \caption{Overall performance and fairness on federated node classification (mean $\pm$ std). We report Overall-F1, Accuracy, Hete-F1, and Hete-min-F1. Best results are in \colorbox[HTML]{DADADA}{\textbf{bold}} and second-best are \underline{underlined}.}
    \vspace{-1pt}
    \resizebox{\textwidth}{!}{
    \begin{tabular}{c|c|c c c c c c c c c}
    \specialrule{1.5pt}{1.5pt}{1.5pt}
    \multicolumn{2}{c|}{\textbf{Description}} 
    & \multicolumn{3}{c}{\textbf{Citation Network}} 
    & \multicolumn{2}{c}{\textbf{Co-author Network}} 
    & \multicolumn{2}{c}{\textbf{Wiki-page Network}}
    & \multicolumn{2}{c}{\textbf{OGB Network}}\\ 
    \cmidrule(lr){1-2} \cmidrule(lr){3-5} \cmidrule(lr){6-7} \cmidrule(lr){8-9} \cmidrule(lr){10-11}
    \multicolumn{2}{c|}{\textbf{Subgraph-FL}} & \textbf{Cora} & \textbf{CiteSeer} & \textbf{PubMed} & \textbf{CS} & \textbf{Physics}
    & \textbf{Chameleon} & \textbf{Squirrel} & \textbf{Ogbn-arxiv} & \textbf{Products}  \\ 
    \midrule

    \multirow{9}{*}{%
  \colorbox[HTML]{E6F7FF}{%
    \parbox[c][3.6cm][c]{2cm}{\centering\textbf{Overall-F1}}%
  }%
}
    & \cellcolor[HTML]{FFF0D5} FedAvg & 
    $60.16_{\scriptstyle \pm 0.51}$ & 
    $57.49_{\scriptstyle \pm 0.22}$ & 
    $70.08_{\scriptstyle \pm 0.36}$ & 
    $80.28_{\scriptstyle \pm 0.57}$ & 
    $84.33_{\scriptstyle \pm 0.83}$ & 
    $58.12_{\scriptstyle \pm 0.25}$ & 
    $50.36_{\scriptstyle \pm 0.27}$ &
    $38.23_{\scriptstyle \pm 0.61}$ &
    $22.31_{\scriptstyle \pm 0.97}$ \\ 
    & \cellcolor[HTML]{FFF0D5} MOON  & 
    $60.67_{\scriptstyle \pm 0.27}$ & 
    $57.31_{\scriptstyle \pm 0.24}$ & 
    $69.84_{\scriptstyle \pm 0.21}$ & 
    $79.28_{\scriptstyle \pm 0.21}$ & 
    $82.44_{\scriptstyle \pm 0.19}$ & 
    $57.85_{\scriptstyle \pm 0.08}$ & 
    $49.91_{\scriptstyle \pm 0.15}$ &
    $38.44_{\scriptstyle \pm 0.15}$ &
    $\underline{23.18_{\scriptstyle \pm 0.57}}$ \\ 
    & \cellcolor[HTML]{FFF7CD} FedSage+  & 
    $55.20_{\scriptstyle \pm 0.35}$ & 
    $58.95_{\scriptstyle \pm 0.26}$ & 
    $70.80_{\scriptstyle \pm 0.41}$ & 
    $80.06_{\scriptstyle \pm 0.28}$ & 
    $83.12_{\scriptstyle \pm 0.39}$ & 
    $61.20_{\scriptstyle \pm 0.75}$ & 
    $50.11_{\scriptstyle \pm 0.22}$ &
    $39.12_{\scriptstyle \pm 0.33}$ &
    $OOM$ \\ 
    & \cellcolor[HTML]{FFF7CD} FedTAD  & 
    $59.38_{\scriptstyle \pm 0.36}$ & 
    $59.08_{\scriptstyle \pm 0.27}$ & 
    $69.52_{\scriptstyle \pm 0.29}$ & 
    $81.06_{\scriptstyle \pm 0.52}$ & 
    $84.12_{\scriptstyle \pm 0.87}$ & 
    $57.55_{\scriptstyle \pm 0.23}$ & 
    $49.13_{\scriptstyle \pm 0.77}$ &
    $39.23_{\scriptstyle \pm 0.32}$ &
    $OOM$ \\ 
    & \cellcolor[HTML]{FFF7CD} AdaFGL  & 
    $58.61_{\scriptstyle \pm 0.32}$ & 
    $58.94_{\scriptstyle \pm 0.55}$ & 
    $70.05_{\scriptstyle \pm 0.26}$ & 
    $80.30_{\scriptstyle \pm 0.38}$ & 
    $83.36_{\scriptstyle \pm 0.92}$ & 
    $58.59_{\scriptstyle \pm 0.41}$ & 
    $50.82_{\scriptstyle \pm 0.39}$ &
    $38.52_{\scriptstyle \pm 0.17}$ & 
    $OOM$ \\ 
    & \cellcolor[HTML]{FFF7CD} FedGTA  & 
    $58.32_{\scriptstyle \pm 0.36}$ & 
    $57.25_{\scriptstyle \pm 0.51}$ & 
    $68.02_{\scriptstyle \pm 0.27}$ & 
    $81.24_{\scriptstyle \pm 0.83}$ & 
    $84.22_{\scriptstyle \pm 0.91}$ & 
    $58.21_{\scriptstyle \pm 0.21}$ & 
    $45.43_{\scriptstyle \pm 0.34}$ &
    $39.23_{\scriptstyle \pm 0.62}$ &
    $OOM$ \\ 
    & \cellcolor[HTML]{F4FFB8} FedSpary  & 
    $50.21_{\scriptstyle \pm 0.34}$ & 
    $57.43_{\scriptstyle \pm 0.22}$ & 
    $74.72_{\scriptstyle \pm 0.36}$ & 
    $80.22_{\scriptstyle \pm 0.57}$ & 
    $84.12_{\scriptstyle \pm 0.83}$ & 
    $58.08_{\scriptstyle \pm 0.25}$ & 
    $53.22_{\scriptstyle \pm 0.27}$ &
    $\underline{39.44_{\scriptstyle \pm 0.61}}$ &
    $OOM$ \\ 
    & \cellcolor[HTML]{F4FFB8} FairFGL  & 
    $\underline{63.87_{\scriptstyle \pm 0.32}}$ & 
    $\underline{61.36_{\scriptstyle \pm 0.36}}$ & $\underline{81.33_{\scriptstyle \pm 0.27}}$ & 
    $\underline{81.08_{\scriptstyle \pm 0.32}}$ & $\underline{84.63_{\scriptstyle \pm 0.19}}$ & $\underline{62.47_{\scriptstyle \pm 0.24}}$ & $\underline{53.75_{\scriptstyle \pm 0.48}}$ &
    $39.24_{\scriptstyle \pm 0.81}$ &
    $OOM$ \\  
    & \cellcolor[HTML]{E6FDD1} BoostFGL & 
    \cellcolor[HTML]{DADADA}$\textbf{66.16}_{\scriptstyle \pm \textbf{0.53}}$ & 
    \cellcolor[HTML]{DADADA}$\textbf{63.85}_{\scriptstyle \pm \textbf{0.31}}$ & 
    \cellcolor[HTML]{DADADA}$\textbf{82.12}_{\scriptstyle \pm \textbf{0.28}}$ & 
    \cellcolor[HTML]{DADADA}$\textbf{89.23}_{\scriptstyle \pm \textbf{0.34}}$ & 
    \cellcolor[HTML]{DADADA}$\textbf{92.09}_{\scriptstyle \pm \textbf{0.14}}$ & 
    \cellcolor[HTML]{DADADA}$\textbf{70.45}_{\scriptstyle \pm \textbf{0.26}}$ & 
    \cellcolor[HTML]{DADADA}$\textbf{61.16}_{\scriptstyle \pm \textbf{0.41}}$ &
    \cellcolor[HTML]{DADADA}$\textbf{43.32}_{\scriptstyle \pm \textbf{0.80}}$ &
    \cellcolor[HTML]{DADADA}$\textbf{29.36}_{\scriptstyle \pm \textbf{0.34}}$ \\ 
    \midrule

    \multirow{9}{*}{%
  \colorbox[HTML]{D6E5FF}{%
    \parbox[c][3.6cm][c]{2cm}{\centering\textbf{Acc}}%
  }%
}
    & \cellcolor[HTML]{FFF0D5} FedAvg & 
    $80.22_{\scriptstyle \pm 0.26}$ & 
    $67.00_{\scriptstyle \pm 0.23}$ & 
    $83.80_{\scriptstyle \pm 0.20}$ & 
    $89.12_{\scriptstyle \pm 0.22}$ & 
    $95.74_{\scriptstyle \pm 0.18}$ & 
    $61.14_{\scriptstyle \pm 0.09}$ & 
    $42.91_{\scriptstyle \pm 0.14}$ &
    $61.16_{\scriptstyle \pm 0.16}$ &
    $46.28_{\scriptstyle \pm 0.32}$ \\ 
    & \cellcolor[HTML]{FFF0D5} MOON  & 
    $80.46_{\scriptstyle \pm 0.72}$ & 
    $67.95_{\scriptstyle \pm 0.42}$ & 
    $83.75_{\scriptstyle \pm 0.12}$ & 
    $88.87_{\scriptstyle \pm 0.12}$ & 
    $94.86_{\scriptstyle \pm 0.91}$ & 
    $60.88_{\scriptstyle \pm 0.80}$ & 
    $43.03_{\scriptstyle \pm 0.51}$ &
    $60.02_{\scriptstyle \pm 0.51}$ &
    $\underline{46.94_{\scriptstyle \pm 0.38}}$ \\ 
    & \cellcolor[HTML]{FFF7CD} FedSage+  & 
    $79.84_{\scriptstyle \pm 0.39}$ & 
    $\underline{71.48_{\scriptstyle \pm 0.46}}$ & 
    $85.06_{\scriptstyle \pm 0.34}$ & 
    $90.43_{\scriptstyle \pm 0.52}$ & 
    $95.26_{\scriptstyle \pm 0.72}$ & 
    $62.49_{\scriptstyle \pm 0.11}$ & 
    $44.46_{\scriptstyle \pm 0.34}$ &
    $60.83_{\scriptstyle \pm 0.67}$ &
    $OOM$ \\ 
    & \cellcolor[HTML]{FFF7CD} FedTAD  & 
    $\underline{81.94_{\scriptstyle \pm 0.90}}$ & 
    $66.61_{\scriptstyle \pm 0.37}$ & 
    $83.64_{\scriptstyle \pm 0.28}$ & 
    $91.73_{\scriptstyle \pm 0.32}$ & 
    $94.84_{\scriptstyle \pm 0.34}$ & 
    $60.26_{\scriptstyle \pm 0.56}$ & 
    $42.63_{\scriptstyle \pm 0.27}$ &
    $61.23_{\scriptstyle \pm 0.95}$ &
    $OOM$ \\  
    & \cellcolor[HTML]{FFF7CD} AdaFGL  & 
    $81.76_{\scriptstyle \pm 0.51}$ & 
    $68.45_{\scriptstyle \pm 0.15}$ & 
    $83.85_{\scriptstyle \pm 0.80}$ & 
    $91.24_{\scriptstyle \pm 0.21}$ & 
    $95.13_{\scriptstyle \pm 0.36}$ & 
    $62.65_{\scriptstyle \pm 0.27}$ & 
    $\underline{45.13_{\scriptstyle \pm 0.93}}$ &
    $60.52_{\scriptstyle \pm 0.19}$ &
    $OOM$ \\ 
    & \cellcolor[HTML]{FFF7CD} FedGTA  & 
    $80.15_{\scriptstyle \pm 0.88}$ & 
    $66.76_{\scriptstyle \pm 0.36}$ & 
    $83.20_{\scriptstyle \pm 0.42}$ & 
    $\underline{92.26_{\scriptstyle \pm 0.42}}$ & 
    $95.24_{\scriptstyle \pm 0.33}$ & 
    $59.69_{\scriptstyle \pm 0.26}$ & 
    $40.54_{\scriptstyle \pm 0.39}$ &
    $59.23_{\scriptstyle \pm 0.30}$ &
    $OOM$ \\  
    & \cellcolor[HTML]{F4FFB8} FedSpary  & 
    $71.22_{\scriptstyle \pm 0.26}$ & 
    $66.94_{\scriptstyle \pm 0.23}$ & 
    $82.26_{\scriptstyle \pm 0.20}$ & 
    $91.02_{\scriptstyle \pm 0.22}$ & 
    $95.35_{\scriptstyle \pm 0.18}$ & 
    $63.42_{\scriptstyle \pm 0.09}$ & 
    $45.03_{\scriptstyle \pm 0.14}$ &
    $62.44_{\scriptstyle \pm 0.16}$ &
    $OOM$ \\ 
    & \cellcolor[HTML]{F4FFB8} FairFGL  & 
    $80.49_{\scriptstyle \pm 0.73}$ &
    $69.13_{\scriptstyle \pm 0.71}$ & 
    $\underline{85.56_{\scriptstyle \pm 0.42}}$ & 
    $90.82_{\scriptstyle \pm 0.48}$ & 
    $\underline{96.03_{\scriptstyle \pm 0.37}}$ & $\underline{64.07_{\scriptstyle \pm 0.53}}$ & 
    $44.42_{\scriptstyle \pm 0.66}$ &
    $\underline{63.24_{\scriptstyle \pm 0.52}}$ &
    $OOM$ \\  
    & \cellcolor[HTML]{E6FDD1} BoostFGL & 
    \cellcolor[HTML]{DADADA}$\textbf{83.02}_{\scriptstyle \pm \textbf{0.43}}$ & 
    \cellcolor[HTML]{DADADA}$\textbf{72.22}_{\scriptstyle \pm \textbf{0.23}}$ & 
    \cellcolor[HTML]{DADADA}$\textbf{87.20}_{\scriptstyle \pm \textbf{0.57}}$ & 
    \cellcolor[HTML]{DADADA}$\textbf{92.44}_{\scriptstyle \pm \textbf{0.73}}$ & 
    \cellcolor[HTML]{DADADA}$\textbf{96.56}_{\scriptstyle \pm \textbf{0.28}}$ & 
    \cellcolor[HTML]{DADADA}$\textbf{65.16}_{\scriptstyle \pm \textbf{0.31}}$ & 
    \cellcolor[HTML]{DADADA}$\textbf{47.52}_{\scriptstyle \pm \textbf{0.26}}$ &
    \cellcolor[HTML]{DADADA}$\textbf{66.32}_{\scriptstyle \pm \textbf{0.62}}$ &
    \cellcolor[HTML]{DADADA}$\textbf{52.14}_{\scriptstyle \pm \textbf{0.28}}$ \\ 

    \specialrule{1.3pt}{2.0pt}{1.0pt}
    \end{tabular}}
    \vspace{-3pt}
    
    \label{table:node_cls}

    \vspace{5pt}

    \resizebox{\textwidth}{!}{
    \begin{tabular}{c|c|c c c c c c c c c}
    \specialrule{1.5pt}{1.5pt}{1.5pt}
    \multicolumn{2}{c|}{\textbf{Description}} 
    & \multicolumn{3}{c}{\textbf{Citation Network}} 
    & \multicolumn{2}{c}{\textbf{Co-author Network}} 
    & \multicolumn{2}{c}{\textbf{Wiki-page Network}}
    & \multicolumn{2}{c}{\textbf{OGB Network}}\\ 
    \cmidrule(lr){1-2} \cmidrule(lr){3-5} \cmidrule(lr){6-7} \cmidrule(lr){8-9} \cmidrule(lr){10-11}
    \multicolumn{2}{c|}{\textbf{Subgraph-FL}} & \textbf{Cora} & \textbf{CiteSeer} & \textbf{PubMed} & \textbf{CS} & \textbf{Physics}
    & \textbf{Chameleon} & \textbf{Squirrel} & \textbf{Ogbn-arxiv} & \textbf{Products}  \\ 
    \midrule

    \multirow{9}{*}{%
  \colorbox[HTML]{F0F5FF}{%
    \parbox[c][3.6cm][c]{2cm}{\centering\textbf{Hete-F1}}%
  }%
}
    & \cellcolor[HTML]{FFF0D5} FedAvg & 
    $45.12_{\scriptstyle \pm 0.38}$ & 
    $41.77_{\scriptstyle \pm 0.91}$ & 
    $53.25_{\scriptstyle \pm 0.82}$ & 
    $53.35_{\scriptstyle \pm 0.28}$ & 
    $56.21_{\scriptstyle \pm 0.49}$ & 
    $49.98_{\scriptstyle \pm 0.51}$ & 
    $42.24_{\scriptstyle \pm 0.93}$ &
    $21.95_{\scriptstyle \pm 0.76}$ &
    $12.41_{\scriptstyle \pm 0.15}$ \\  
    & \cellcolor[HTML]{FFF0D5} MOON  & 
    $45.24_{\scriptstyle \pm 0.42}$ & 
    $41.45_{\scriptstyle \pm 0.22}$ & 
    $53.11_{\scriptstyle \pm 0.33}$ & 
    $53.31_{\scriptstyle \pm 0.54}$ & 
    $56.28_{\scriptstyle \pm 0.38}$ & 
    $50.12_{\scriptstyle \pm 0.25}$ & 
    $42.36_{\scriptstyle \pm 0.92}$ &
    $22.81_{\scriptstyle \pm 0.26}$ &
    $\underline{13.57_{\scriptstyle \pm 0.15}}$ \\ 
    & \cellcolor[HTML]{FFF7CD} FedSage+  & 
    $45.32_{\scriptstyle \pm 0.72}$ & 
    $46.43_{\scriptstyle \pm 0.46}$ & 
    $61.53_{\scriptstyle \pm 0.28}$ & 
    $61.23_{\scriptstyle \pm 0.42}$ & 
    $64.18_{\scriptstyle \pm 0.18}$ & 
    $58.28_{\scriptstyle \pm 0.12}$ & 
    $50.36_{\scriptstyle \pm 0.14}$ &
    $30.23_{\scriptstyle \pm 0.19}$ &
    $OOM$ \\ 
    & \cellcolor[HTML]{FFF7CD} FedTAD  & 
    $44.28_{\scriptstyle \pm 0.33}$ & 
    $43.02_{\scriptstyle \pm 0.28}$ & 
    $52.84_{\scriptstyle \pm 0.45}$ & 
    $52.91_{\scriptstyle \pm 0.77}$ & 
    $56.02_{\scriptstyle \pm 0.31}$ & 
    $50.29_{\scriptstyle \pm 0.55}$ & 
    $42.78_{\scriptstyle \pm 0.67}$ &
    $21.54_{\scriptstyle \pm 0.26}$ &
    $OOM$ \\ 
    & \cellcolor[HTML]{FFF7CD} AdaFGL  & 
    $44.39_{\scriptstyle \pm 0.23}$ & 
    $43.10_{\scriptstyle \pm 0.35}$ & 
    $54.41_{\scriptstyle \pm 0.36}$ & 
    $54.28_{\scriptstyle \pm 0.33}$ & 
    $57.24_{\scriptstyle \pm 0.85}$ & 
    $51.55_{\scriptstyle \pm 0.24}$ & 
    $43.95_{\scriptstyle \pm 0.31}$ &
    $23.11_{\scriptstyle \pm 0.24}$ &
    $OOM$ \\ 
    & \cellcolor[HTML]{FFF7CD} FedGTA  & 
    $44.28_{\scriptstyle \pm 0.43}$ & 
    $43.02_{\scriptstyle \pm 0.25}$ & 
    $51.54_{\scriptstyle \pm 0.66}$ & 
    $51.37_{\scriptstyle \pm 0.39}$ & 
    $54.28_{\scriptstyle \pm 0.28}$ & 
    $58.32_{\scriptstyle \pm 0.42}$ & 
    $50.28_{\scriptstyle \pm 0.17}$ &
    $20.24_{\scriptstyle \pm 0.77}$ &
    $OOM$ \\ 
    & \cellcolor[HTML]{F4FFB8} FedSpary  & 
    $45.36_{\scriptstyle \pm 0.31}$ & 
    $43.35_{\scriptstyle \pm 0.26}$ & 
    $60.81_{\scriptstyle \pm 0.27}$ & 
    $61.88_{\scriptstyle \pm 0.37}$ & 
    $65.02_{\scriptstyle \pm 0.74}$ & 
    $60.12_{\scriptstyle \pm 0.38}$ & 
    $52.15_{\scriptstyle \pm 0.55}$ &
    $29.41_{\scriptstyle \pm 0.82}$ &
    $OOM$ \\ 
    & \cellcolor[HTML]{F4FFB8} FairFGL  & 
    $\underline{47.56_{\scriptstyle \pm 0.29}}$ & $\underline{48.45_{\scriptstyle \pm 0.48}}$ & $\underline{65.66_{\scriptstyle \pm 0.36}}$ & 
    $\underline{65.12_{\scriptstyle \pm 0.26}}$ & $\underline{68.24_{\scriptstyle \pm 0.47}}$ & $\underline{62.12_{\scriptstyle \pm 0.31}}$ & $\underline{54.36_{\scriptstyle \pm 0.27}}$ &
    $\underline{34.36_{\scriptstyle \pm 0.51}}$ &
    $OOM$ \\  
    & \cellcolor[HTML]{E6FDD1} BoostFGL & 
    \cellcolor[HTML]{DADADA}$\textbf{52.06}_{\scriptstyle \pm \textbf{0.25}}$ & 
    \cellcolor[HTML]{DADADA}$\textbf{54.42}_{\scriptstyle \pm \textbf{0.73}}$ & 
    \cellcolor[HTML]{DADADA}$\textbf{68.42}_{\scriptstyle \pm \textbf{0.68}}$ & 
    \cellcolor[HTML]{DADADA}$\textbf{71.33}_{\scriptstyle \pm \textbf{0.31}}$ & 
    \cellcolor[HTML]{DADADA}$\textbf{75.12}_{\scriptstyle \pm \textbf{0.57}}$ & 
    \cellcolor[HTML]{DADADA}$\textbf{73.24}_{\scriptstyle \pm \textbf{0.36}}$ & 
    \cellcolor[HTML]{DADADA}$\textbf{57.32}_{\scriptstyle \pm \textbf{0.51}}$ &
    \cellcolor[HTML]{DADADA}$\textbf{37.12}_{\scriptstyle \pm \textbf{0.24}}$ &
    \cellcolor[HTML]{DADADA}$\textbf{18.33}_{\scriptstyle \pm \textbf{0.34}}$ \\ 
    \midrule

    \multirow{9}{*}{%
  \colorbox[HTML]{E3EEFA}{%
    \parbox[c][3.6cm][c]{2cm}{\centering\textbf{Hete-min-F1}}%
  }%
} 
    & \cellcolor[HTML]{FFF0D5} FedAvg & 
    $30.39_{\scriptstyle \pm 0.38}$ & 
    $25.33_{\scriptstyle \pm 0.47}$ & 
    $45.34_{\scriptstyle \pm 0.95}$ & 
    $47.31_{\scriptstyle \pm 0.36}$ & 
    $50.12_{\scriptstyle \pm 0.31}$ & 
    $35.64_{\scriptstyle \pm 0.57}$ & 
    $27.43_{\scriptstyle \pm 0.68}$ &
    $12.48_{\scriptstyle \pm 0.27}$ &
    $5.67_{\scriptstyle \pm 0.31}$ \\  
    & \cellcolor[HTML]{FFF0D5} MOON  & 
    $31.37_{\scriptstyle \pm 0.51}$ & 
    $25.66_{\scriptstyle \pm 0.27}$ & 
    $45.76_{\scriptstyle \pm 0.35}$ & 
    $48.28_{\scriptstyle \pm 0.17}$ & 
    $51.24_{\scriptstyle \pm 0.85}$ & 
    $35.92_{\scriptstyle \pm 0.30}$ & 
    $27.92_{\scriptstyle \pm 0.84}$ &
    $12.54_{\scriptstyle \pm 0.36}$ &
    \underline{$6.10_{\scriptstyle \pm 0.19}$} \\ 
    & \cellcolor[HTML]{FFF7CD} FedSage+  & 
    $22.71_{\scriptstyle \pm 0.27}$ & 
    $27.20_{\scriptstyle \pm 0.31}$ & 
    $58.74_{\scriptstyle \pm 0.73}$ & 
    $39.33_{\scriptstyle \pm 0.86}$ & 
    $42.14_{\scriptstyle \pm 0.31}$ & 
    $29.86_{\scriptstyle \pm 0.59}$ & 
    $23.12_{\scriptstyle \pm 0.26}$ &
    $20.28_{\scriptstyle \pm 0.50}$ &
    $OOM$ \\ 
    & \cellcolor[HTML]{FFF7CD} FedTAD  & 
    $27.74_{\scriptstyle \pm 0.35}$ & 
    $25.19_{\scriptstyle \pm 0.26}$ & 
    $44.37_{\scriptstyle \pm 0.33}$ & 
    $44.21_{\scriptstyle \pm 0.72}$ & 
    $48.32_{\scriptstyle \pm 0.33}$ & 
    $33.12_{\scriptstyle \pm 0.21}$ & 
    $25.48_{\scriptstyle \pm 0.09}$ &
    $12.45_{\scriptstyle \pm 0.64}$ &
    $OOM$ \\
    & \cellcolor[HTML]{FFF7CD} AdaFGL  & 
    $31.54_{\scriptstyle \pm 0.25}$ & 
    $28.72_{\scriptstyle \pm 0.32}$ & 
    $47.48_{\scriptstyle \pm 0.87}$ & 
    $46.98_{\scriptstyle \pm 0.52}$ & 
    $49.28_{\scriptstyle \pm 0.24}$ & 
    $33.69_{\scriptstyle \pm 0.12}$ & 
    $25.96_{\scriptstyle \pm 0.22}$ &
    $15.12_{\scriptstyle \pm 0.26}$ &
    $OOM$ \\ 
    & \cellcolor[HTML]{FFF7CD} FedGTA  & 
    $29.86_{\scriptstyle \pm 0.19}$ & 
    $24.38_{\scriptstyle \pm 0.81}$ & 
    $41.95_{\scriptstyle \pm 0.09}$ & 
    $37.24_{\scriptstyle \pm 0.82}$ & 
    $40.42_{\scriptstyle \pm 0.10}$ & 
    $24.28_{\scriptstyle \pm 0.30}$ & 
    $26.24_{\scriptstyle \pm 0.06}$ &
    $10.21_{\scriptstyle \pm 0.21}$ &
    $OOM$ \\ 
    & \cellcolor[HTML]{F4FFB8} FedSpary  & 
    $26.56_{\scriptstyle \pm 0.26}$ & 
    $26.51_{\scriptstyle \pm 0.03}$ & 
    $55.86_{\scriptstyle \pm 0.36}$ & 
    $43.22_{\scriptstyle \pm 0.28}$ & 
    $48.34_{\scriptstyle \pm 0.94}$ & 
    $28.27_{\scriptstyle \pm 0.32}$ & 
    $30.97_{\scriptstyle \pm 0.17}$ &
    $21.69_{\scriptstyle \pm 0.71}$ &
    $OOM$ \\ 
    & \cellcolor[HTML]{F4FFB8} FairFGL  & 
    \underline{$37.10_{\scriptstyle \pm 0.31}$} & 
    \underline{$34.99_{\scriptstyle \pm 0.27}$} & 
    \underline{$61.04_{\scriptstyle \pm 0.88}$} & 
    \underline{$54.21_{\scriptstyle \pm 0.33}$} & 
    \underline{$56.28_{\scriptstyle \pm 0.26}$} & 
    \underline{$39.47_{\scriptstyle \pm 0.51}$} & 
    \underline{$32.31_{\scriptstyle \pm 0.24}$} &
    \underline{$28.14_{\scriptstyle \pm 0.92}$} & 
    $OOM$ \\ 
    & \cellcolor[HTML]{E6FDD1} BoostFGL & 
    \cellcolor[HTML]{DADADA}$\textbf{41.25}_{\scriptstyle \pm \textbf{0.28}}$ & 
    \cellcolor[HTML]{DADADA}$\textbf{38.32}_{\scriptstyle \pm \textbf{0.36}}$ & 
    \cellcolor[HTML]{DADADA}$\textbf{64.28}_{\scriptstyle \pm \textbf{0.54}}$ & 
    \cellcolor[HTML]{DADADA}$\textbf{68.51}_{\scriptstyle \pm \textbf{0.37}}$ & 
    \cellcolor[HTML]{DADADA}$\textbf{72.14}_{\scriptstyle \pm \textbf{0.36}}$ & 
    \cellcolor[HTML]{DADADA}$\textbf{50.02}_{\scriptstyle \pm \textbf{0.21}}$ & 
    \cellcolor[HTML]{DADADA}$\textbf{35.12}_{\scriptstyle \pm \textbf{0.98}}$ &
    \cellcolor[HTML]{DADADA}$\textbf{33.13}_{\scriptstyle \pm \textbf{0.21}}$ &
    \cellcolor[HTML]{DADADA}$\textbf{11.25}_{\scriptstyle \pm \textbf{0.26}}$ \\

    \specialrule{1.3pt}{2.0pt}{1.0pt}
    \end{tabular}}

    \end{table*}

\begin{table*}[htbp]
    \centering
    \caption{Ablation study on Cora and Chameleon datasets. Each variant removes one key boosting mechanism to validate its contribution. Bold values indicate the best performance in each column.}
    \vspace{-0pt}
    \resizebox{\textwidth}{!}{
    \begin{tabular}{c|c|cccc|cccc}
    \specialrule{1.5pt}{1.5pt}{1.5pt}
    \multirow{2}{*}{\textbf{Dataset}} &
    \multirow{2}{*}{\textbf{Variant}} &
    \multicolumn{4}{c|}{\textbf{Cora}} &
    \multicolumn{4}{c}{\textbf{Chameleon}} \\
    \cmidrule(lr){3-6} \cmidrule(lr){7-10}
    & & \textbf{Overall-F1} & \textbf{Acc} & \textbf{hete-F1} & \textbf{hete-min-F1} & \textbf{Overall-F1} & \textbf{Acc} & \textbf{hete-F1} & \textbf{hete-min-F1} \\
    \midrule

    \multirow{4}{*}{\rotatebox{90}{\textbf{Ablation}}}
    & Full BoostFGL
    & \cellcolor[HTML]{DADADA}$\textbf{66.16}_{\scriptstyle \pm \textbf{0.53}}$
    & \cellcolor[HTML]{DADADA}$\textbf{83.02}_{\scriptstyle \pm \textbf{0.43}}$
    & \cellcolor[HTML]{DADADA}$\textbf{52.06}_{\scriptstyle \pm \textbf{0.25}}$
    & \cellcolor[HTML]{DADADA}$\textbf{41.25}_{\scriptstyle \pm \textbf{0.28}}$
    & \cellcolor[HTML]{DADADA}$\textbf{70.45}_{\scriptstyle \pm \textbf{0.26}}$
    & \cellcolor[HTML]{DADADA}$\textbf{65.16}_{\scriptstyle \pm \textbf{0.31}}$
    & \cellcolor[HTML]{DADADA}$\textbf{73.24}_{\scriptstyle \pm \textbf{0.36}}$
    & \cellcolor[HTML]{DADADA}$\textbf{50.02}_{\scriptstyle \pm \textbf{0.21}}$ \\

    & w/o Node Boosting
    & \underline{$62.49_{\scriptstyle \pm 0.65}$}
    & \underline{$80.25_{\scriptstyle \pm 0.52}$}
    & \underline{$49.40_{\scriptstyle \pm 0.31}$}
    & \underline{$37.97_{\scriptstyle \pm 0.42}$}
    & \underline{$67.89_{\scriptstyle \pm 0.31}$}
    & \underline{$62.59_{\scriptstyle \pm 0.37}$}
    & \underline{$61.11_{\scriptstyle \pm 0.43}$}
    & \underline{$44.81_{\scriptstyle \pm 0.29}$} \\

    & w/o Topology Boosting
    & $60.38_{\scriptstyle \pm 0.72}$
    & $78.36_{\scriptstyle \pm 0.61}$
    & $47.29_{\scriptstyle \pm 0.40}$
    & $33.16_{\scriptstyle \pm 0.60}$
    & $65.67_{\scriptstyle \pm 0.45}$
    & $60.35_{\scriptstyle \pm 0.49}$
    & $58.80_{\scriptstyle \pm 0.58}$
    & $43.60_{\scriptstyle \pm 0.54}$ \\

    & w/o Model Boosting
    & $61.82_{\scriptstyle \pm 0.69}$
    & $79.47_{\scriptstyle \pm 0.56}$
    & $48.51_{\scriptstyle \pm 0.36}$
    & $33.88_{\scriptstyle \pm 0.49}$
    & $67.02_{\scriptstyle \pm 0.38}$
    & $61.74_{\scriptstyle \pm 0.44}$
    & $60.19_{\scriptstyle \pm 0.50}$
    & $44.35_{\scriptstyle \pm 0.38}$ \\

    \specialrule{1.3pt}{2.0pt}{1.0pt}
    \end{tabular}}
    \vspace{-3pt}
    \label{table:ablation}
\end{table*}

\section{Experiments}
In this section, we provide a comprehensive empirical evaluation of BoostFGL around the following research questions: \textbf{Q1} – Does BoostFGL improve overall accuracy and fairness over strong federated graph learning baselines across benchmarks? \textbf{Q2} – What is the individual contribution of each boosting module in BoostFGL? \textbf{Q3} – How robust is BoostFGL to hyper-parameter choices and training dynamics? \textbf{Q4} – What communication and computation efficiency does BoostFGL achieve compared to baselines? \textbf{Q5} – Can BoostFGL be combined with standard privacy mechanisms while maintaining competitive performance?

\vspace{-8pt}

\subsection{Experimental Setup}
We evaluate BoostFGL under standard subgraph-FL protocols on nine widely used benchmarks spanning diverse graph scales and heterogeneity patterns, including citation, co-author, wiki-page, and large-scale OGB graphs.
For simulation, we follow common FGL practice and partition each graph into structurally coherent client subgraphs via the Louvain algorithm, treating each subgraph as a federated client.
We compare representative baselines from three families—generic FL, FGL methods for graph non-IIDness/personalization, and fairness-centered FGL—and report both utility and fairness metrics.
Specifically, we use Overall-F1 and Accuracy for global performance, together with Hete-F1 and Hete-min-F1 to quantify disparities on heterogeneous node groups.
Full details on datasets, baselines, metrics, and simulation are deferred to Appendix~\ref{app:exp_details}.

\subsection{Overall Performance}
\label{sec:overall_performance}

To answer \textbf{Q1}, we systematically compare BoostFGL with all baselines across the above datasets.
As shown in Table~\ref{table:node_cls}, BoostFGL consistently achieves the best or second-best results across all evaluation metrics, demonstrating strong effectiveness in simultaneously improving predictive performance and fairness.
In particular, BoostFGL yields substantial gains on fairness-oriented metrics while maintaining or even surpassing the performance of accuracy-driven baselines.
Moreover, on large-scale OGB datasets, BoostFGL exhibits superior scalability and robustness: while several competing methods encounter OOM issues on the \textit{products} dataset, BoostFGL remains stable and achieves strong performance, highlighting its practical applicability to real-world large-scale FGL scenarios.
Overall, BoostFGL improves fairness metrics by \textbf{9.6\%} and overall performance by \textbf{8.5\%} compared to the strongest competing baselines.

\vspace{-7pt}

\subsection{Ablation Study}
To answer \textbf{Q2}, we conduct ablation studies to isolate the individual contribution of each boosting module in BoostFGL.
As shown in Table~\ref{table:ablation}, the complete BoostFGL consistently achieves the best performance, whereas removing any single module leads to clear and consistent degradation in both overall performance and fairness.
Specifically, Node Boosting primarily contributes to improved performance on minority and low-frequency nodes, Topology Boosting plays a dominant role in enhancing fairness under heterophilous graph structures, and Model Boosting provides stable and complementary gains for both accuracy and fairness.

\vspace{-7pt}

\subsection{Robustness Analysis}
To answer \textbf{Q3}, we evaluate robustness from two perspectives: hyperparameter sensitivity and training dynamics.
The heatmap over $\tau$ and the edge budget exhibits a broad high-performing region, indicating that performance is stable under a wide range of graph-editing configurations.
Similarly, varying $\lambda$ leads to only modest changes on most datasets, with larger values becoming less stable on structurally challenging graphs; we therefore adopt $\lambda{=}0.5$ as a robust default.
In addition, the convergence curves in Fig.~\ref{fig:convergence} show that our method consistently attains higher Overall-F1 and converges faster than FedAvg, FedSage+, and FairFGL across communication rounds.

\vspace{-7pt}

\begin{figure}[t]
  \centering
  \includegraphics[width=\linewidth]{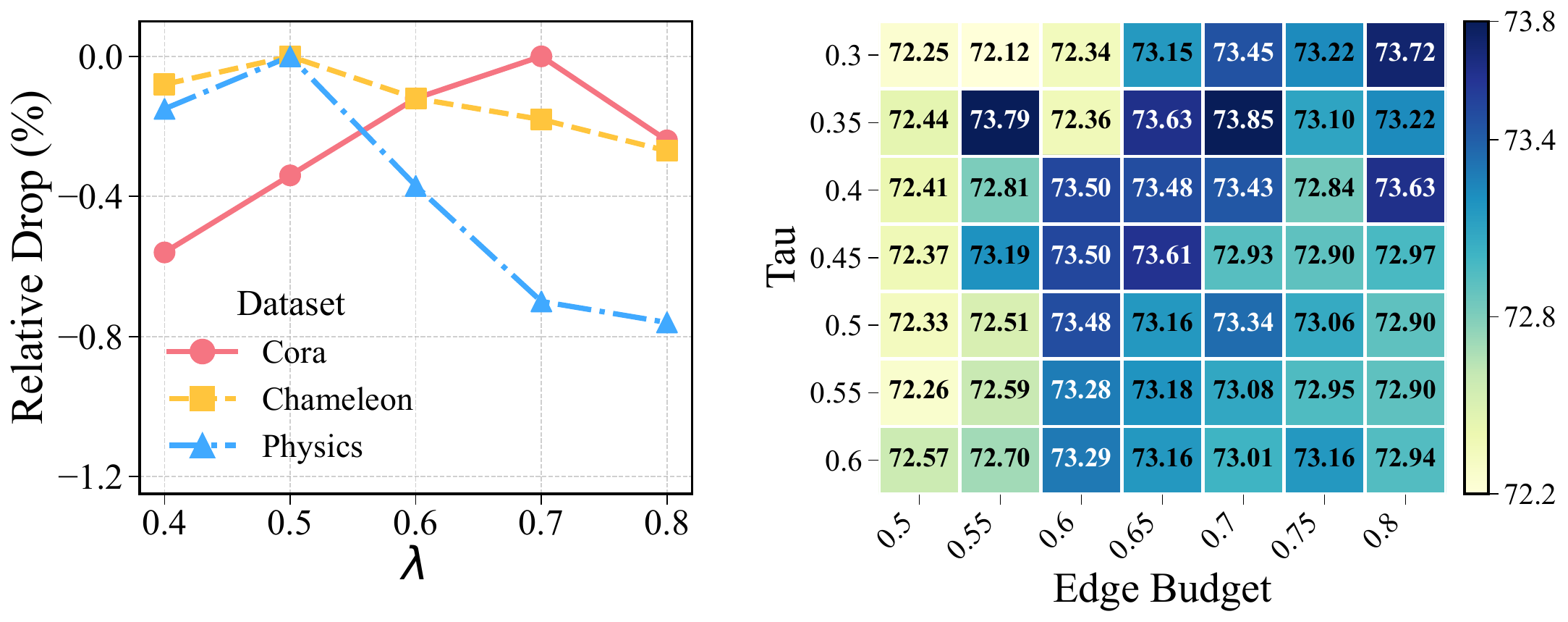}
  \caption{\textbf{Hyperparameter sensitivity.}
\textit{Left:} Relative F1 drop under different node-boosting strengths $\lambda$ on representative datasets.
\textit{Right:} Heatmap over topology- and model-boosting hyperparameters $(\tau,\text{edge budget})$.}
  \label{fig:ha}
\end{figure}

\subsection{Efficiency analysis}

\begin{figure}[t]
  \centering
  \begin{minipage}[t]{0.48\linewidth}
    \centering
    \includegraphics[width=\linewidth]{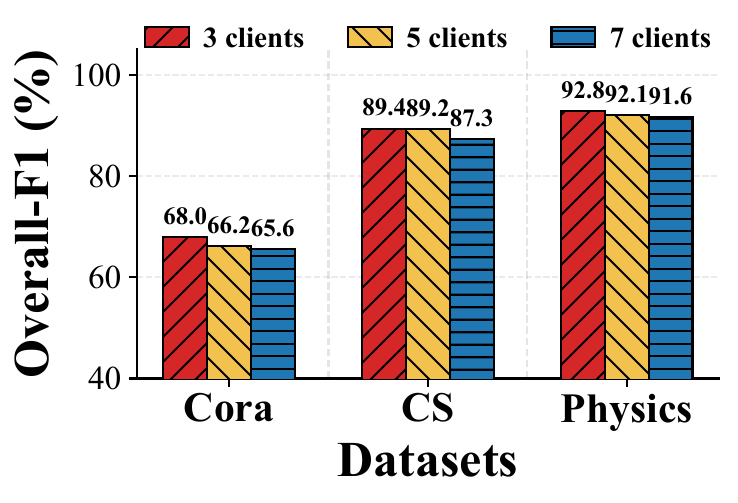}
    \label{fig:cl}
  \end{minipage}
  \hfill
  \begin{minipage}[t]{0.48\linewidth}
    \centering
    \includegraphics[width=\linewidth]{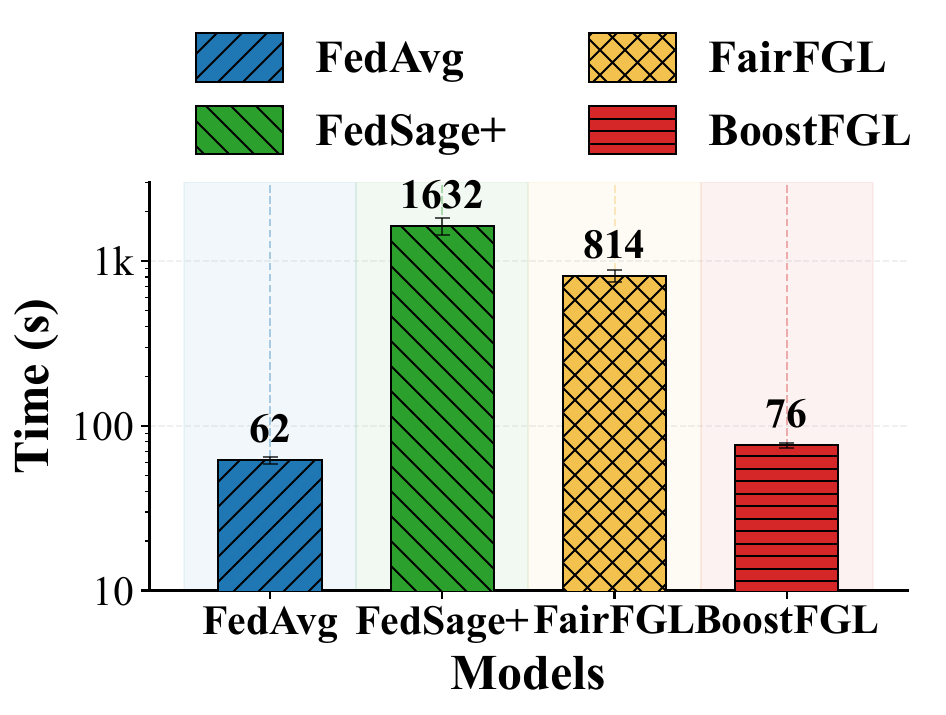}
    \label{fig:ti}
  \end{minipage}
  \caption{\textbf{Scalability and efficiency of BoostFGL.}
BoostFGL maintains stable Overall-F1 when increasing the number of clients and achieves competitive end-to-end runtime compared with representative baselines.}
  \label{fig:cl_ti}
\end{figure}

To answer \textbf{Q4}, we evaluate the scalability and runtime overhead of BoostFGL.
Fig.~\ref{fig:cl_ti} (left) shows that BoostFGL remains stable as the number of clients increases, indicating robustness to more fragmented subgraphs.
Fig.~\ref{fig:cl_ti} (right) reports end-to-end wall-clock time: BoostFGL stays close to lightweight baselines and is markedly faster than heavier FGL methods.
This efficiency is due to its lightweight design—client-side modules are simple local reweighting, and the server-side module only adjusts aggregation using a few statistics without changing the protocol.

\vspace{-7pt}

\subsection{Privacy analysis}

\begin{figure}[t]
  \centering
  \includegraphics[width=\linewidth]{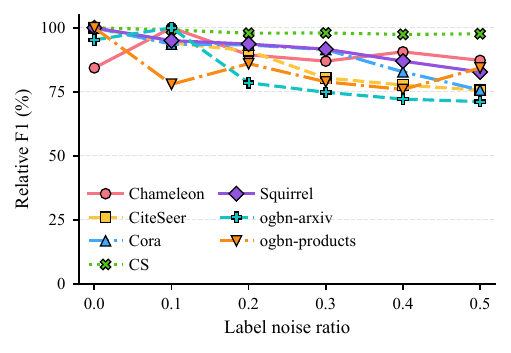}
  \caption{\textbf{DP-noise robustness.} Relative Overall-F1 of BoostFGL under Gaussian noise injected into propagation to simulate tighter DP privacy budgets.}

  \label{fig:noise}
\end{figure}

To answer \textbf{Q5}, we show that BoostFGL can be combined with standard privacy mechanisms without changing the subgraph-FL protocol: clients keep raw data local and only upload model updates, making it compatible with secure aggregation and DP-FL.
Secure aggregation applies directly since the server only reweights aggregated updates.
For DP, BoostFGL can plug into DP-FedAvg by clipping and adding Gaussian noise to $\Delta\theta_m$; we further simulate tighter DP budgets by injecting DP-style noise into propagation and report the resulting relative performance in Fig.~\ref{fig:noise}.

\section{Conclusion}

In this paper, we revisit FGL from a fairness perspective and show that node-level disparities are often \emph{pipeline-induced}, arising from three coupled sources: label skew in local optimization, topology confounding during message propagation, and aggregation dilution across heterogeneous clients. To address these issues in a unified and protocol-compatible manner, we propose \textbf{BoostFGL}, a boosting-style framework that introduces lightweight, modular corrections at three stages of the client--server pipeline. Specifically, BoostFGL amplifies learning signals for systematically under-served nodes, recalibrates propagation over structurally unreliable neighborhoods, and performs difficulty- and reliability-aware aggregation to preserve fairness-critical updates, while remaining compatible with standard backbones, federated protocols, and privacy mechanisms. Extensive experiments on nine datasets demonstrate that BoostFGL improves fairness while maintaining competitive overall performance, with robust hyperparameter behavior and favorable efficiency. More broadly, our results suggest a practical design principle for fairness-aware FGL: unfairness should be diagnosed with stage-specific signals and mitigated at the stage where it arises, rather than being addressed only through a global objective.

\bibliography{example_paper}
\bibliographystyle{icml2025}

\newpage
\appendix
\onecolumn
\section{Experiment Details}
\label{app:exp_details}

This appendix provides complete experimental details for reproducibility, including computation resources,
evaluation metrics (utility and fairness), dataset statistics, federated simulation protocol, and default hyper-parameter settings.

\subsection{Computation Resource}
\label{app:resource}

All experiments are conducted on the same Linux server under an identical hardware/software environment for all compared methods.
We report end-to-end wall-clock runtime measured on this machine for all methods in Fig.~\ref{fig:cl_ti}.

\begin{itemize}
    \item \textbf{CPU:} 2 $\times$ Intel(R) Xeon(R) Gold 6240 @ 2.60GHz (36 cores per socket; 72 threads total).
    \item \textbf{RAM:} 251\,GB (approximately 216\,GB available during experiments).
    \item \textbf{GPU:} 4 $\times$ NVIDIA A100 (40\,GB memory per GPU).
    \item \textbf{Software:} Python 3.11.10; PyTorch 2.1.0 with CUDA 11.8.
\end{itemize}

We adopt a multi-GPU training setup where each simulated client is pinned to a dedicated GPU during federated optimization, enabling parallel execution across clients under the client--client (C--C) communication paradigm.
We define \textbf{OOM} (out-of-memory) as the case where a method exceeds the 40\,GB memory capacity of a single A100 GPU, and \textbf{OOT} (out-of-time) as the case where the end-to-end training time exceeds 8 hours on a single A100 GPU.
These constraints reflect practical resource limits when scaling graph learning to large datasets (e.g., \textsc{ogbn-products}).

\subsection{Evaluation Metrics}
\label{app:metrics}

Let $\mathcal{V}^{\text{test}}$ denote the global test node set and $C$ be the number of classes.
For any evaluated node subset $\mathcal{S}\subseteq \mathcal{V}^{\text{test}}$, let $\hat y_v$ be the predicted label of node $v$ and $y_v$ be its ground-truth label.

\paragraph{Utility metrics.}
We report \textbf{Accuracy} and \textbf{Overall-F1} to measure predictive utility on $\mathcal{V}^{\text{test}}$.

\textbf{Accuracy.}
\begin{equation}
\label{eq:acc}
\mathrm{Acc}(\mathcal{S})
:= \frac{1}{|\mathcal{S}|}\sum_{v\in\mathcal{S}} \mathbb{I}\big[\hat y_v = y_v\big].
\end{equation}
\emph{Why Accuracy:} it captures the overall correctness rate and is the most commonly reported metric for semi-supervised node classification.

\textbf{Overall-F1.}
For class $c\in\{1,\dots,C\}$, define precision and recall on subset $\mathcal{S}$ as
\begin{equation}
\label{eq:prec}
\mathrm{Prec}_c(\mathcal{S})
:= \frac{\sum_{v\in\mathcal{S}}\mathbb{I}[\hat y_v=c]\mathbb{I}[y_v=c]}
{\sum_{v\in\mathcal{S}}\mathbb{I}[\hat y_v=c] + \epsilon},
\end{equation}
\begin{equation}
\label{eq:rec}
\mathrm{Rec}_c(\mathcal{S})
:= \frac{\sum_{v\in\mathcal{S}}\mathbb{I}[\hat y_v=c]\mathbb{I}[y_v=c]}
{\sum_{v\in\mathcal{S}}\mathbb{I}[y_v=c] + \epsilon},
\end{equation}
and the per-class F1 score as
\begin{equation}
\label{eq:f1c}
\mathrm{F1}_c(\mathcal{S})
:= \frac{2\,\mathrm{Prec}_c(\mathcal{S})\,\mathrm{Rec}_c(\mathcal{S})}
{\mathrm{Prec}_c(\mathcal{S})+\mathrm{Rec}_c(\mathcal{S})+\epsilon}.
\end{equation}
Then Overall-F1 on $\mathcal{S}$ is
\begin{equation}
\label{eq:overall_f1}
\mathrm{OverallF1}(\mathcal{S})
:= \frac{1}{C}\sum_{c=1}^{C}\mathrm{F1}_c(\mathcal{S}).
\end{equation}
We set $\epsilon$ to a small constant to avoid division by zero when a class has no predicted/true instances in $\mathcal{S}$.

\textbf{Overall-F1.}
\begin{equation}
\label{eq:overall_f1}
\mathrm{Overall\text{-}F1}
:= \mathrm{OverallF1}(\mathcal{V}^{\text{test}}).
\end{equation}
\emph{Why Overall-F1 / Overall-F1:} compared with Accuracy, Overall-F1 weights classes uniformly and is less dominated by frequent classes, which is critical under label-imbalanced graph benchmarks and for fairness-aware evaluation.

\paragraph{Fairness metrics.}
Following FairFGL~\citep{fairfgl}, we evaluate group-wise Overall-F1 on structurally disadvantaged node subsets.
Let $\mathcal{V}_{\hete}$ be the heterophilous test-node set and $\mathcal{V}_{\hete\text{-}\min}$ be the intersection of heterophilous and minority test nodes.
Their constructions are defined in Appendix~\ref{sec:fairness-challenge}.

\textbf{Hete-F1.}
\begin{equation}
\label{eq:hete_f1}
\mathrm{Hete\text{-}F1}
:= \mathrm{OverallF1}(\mathcal{V}_{\hete}).
\end{equation}

\textbf{Hete-min-F1.}
\begin{equation}
\label{eq:hete_min_f1}
\mathrm{Hete\text{-}min\text{-}F1}
:= \mathrm{OverallF1}(\mathcal{V}_{\hete\text{-}\min}).
\end{equation}

\emph{Why these fairness metrics:}
our goal is to diagnose and mitigate performance degradation on structurally disadvantaged nodes under semi-supervised graph learning.
Heterophilous nodes are harder because their local neighborhoods provide less label-consistent evidence, while minority nodes are harder due to limited labeled supervision.
Overall-F1 on $\mathcal{V}_{\hete}$ and $\mathcal{V}_{\hete\text{-}\min}$ directly measures whether a method improves prediction quality for these disadvantaged groups, which can be hidden by aggregate utility metrics.
We prefer group-wise performance metrics over constraint-based notions (e.g., equalized odds) because the task is node classification on graphs with structure-dependent difficulty, and the key failure mode we target is \emph{group-specific utility collapse} rather than enforcing outcome parity.

\subsection{Fairness Challenge and Group Construction}
\label{sec:fairness-challenge}

This section specifies how we construct the disadvantaged node groups used by
Hete-F1 and Hete-min-F1 (Appendix~\ref{app:metrics}), following FairFGL~\citep{fairfgl}.

\paragraph{Minority group.}
Let $n_c$ be the number of labeled nodes in class $c$ (counted on the global labeled set under each split/seed).
We sort classes increasingly by $n_c$ and define the \emph{minority classes} as the smallest set whose cumulative labeled mass
reaches a ratio $q\in(0,1)$:
\begin{equation}
\label{eq:minority_def}
\mathcal{C}_{\min}(q)
:= \arg\min_{\mathcal{C}\subseteq\{1,\dots,C\}}
\Big\{|\mathcal{C}|:\ \sum_{c\in\mathcal{C}} n_c \ge q\sum_{c=1}^C n_c\Big\},
\qquad
\mathcal{V}_{\min}
:= \{v\in\mathcal{V}^{\text{test}}:\ y_v\in\mathcal{C}_{\min}(q)\}.
\end{equation}
Unless otherwise stated, we use the default $q$ in the main paper and vary $q$ in Sec.~\ref{sec:appendix_empirical_minority}.

\paragraph{Heterophilous group.}
For a test node $v$ with label $y_v$, define its node-level homophily as
\begin{equation}
\label{eq:node_homophily}
\phi(v)
:= \frac{1}{|\mathcal{N}(v)|}\sum_{u\in\mathcal{N}(v)} \mathbb{I}[y_u = y_v].
\end{equation}
Following FairFGL~\citep{fairfgl}, we define heterophilous test nodes as those with low homophily:
\begin{equation}
\label{eq:hete_set}
\mathcal{V}_{\hete}
:= \{v\in\mathcal{V}^{\text{test}}:\ \phi(v) \le \tau_h\},
\end{equation}
where $\tau_h$ is a fixed threshold (default $\tau_h=0.5$ unless otherwise specified).

\paragraph{Intersection group.}
Finally, we define
\begin{equation}
\label{eq:hete_min_set}
\mathcal{V}_{\hete\text{-}\min} := \mathcal{V}_{\hete}\cap \mathcal{V}_{\min}.
\end{equation}

\subsection{Datasets \& Simulation Method}
\label{app:data_partition}

\paragraph{Datasets.}
We conduct experiments on nine widely-used datasets spanning diverse graph scales and heterogeneity patterns, including three citation networks (Cora, CiteSeer, PubMed~\cite{yang2016revisiting}), two co-author networks (CS, Physics~\cite{shchur2018pitfalls}), two wiki-page networks (Chameleon, Squirrel~\cite{rozemberczki2021multi}), and two large-scale OGB networks (ogbn-arxiv, ogbn-products~\cite{hu2020open}).
Dataset statistics and split ratios are summarized in Table~\ref{tab:dataset_statistics}.

\paragraph{Data splits.}
We follow the train/validation/test split ratios in Table~\ref{tab:dataset_statistics} for each dataset.
All compared methods share identical splits under each random seed.

\paragraph{Subgraph-FL simulation via Louvain partition.}
Following standard subgraph-FL settings~\citep{zhang2021subgraph}, we partition each graph into structurally coherent subgraphs using the Louvain community detection algorithm~\citep{blondel2008fast}.
Louvain greedily optimizes modularity by iteratively (i) moving nodes to neighboring communities that increase modularity and
(ii) collapsing discovered communities to form a coarsened graph, producing a hierarchical community structure.
We treat each detected community as one federated client and construct local subgraphs $\{\mathcal{G}_k\}_{k=1}^{K}$ with disjoint node sets.
Unless otherwise stated, we use $K=5$ clients for the main results, and vary $K\in\{3,5,7\}$ in the scalability study (Fig.~\ref{fig:cl_ti}).

\begin{table}[t]
\centering
\caption{Statistical information of Subgraph-FL datasets used in this paper.}
\resizebox{\linewidth}{!}{
\begin{tabular}{lrrrrcl}
\toprule
\textbf{Dataset} & \textbf{Nodes} & \textbf{Edges} & \textbf{Features} & \textbf{Classes} & \textbf{Train/Val/Test} & \textbf{Description} \\
\midrule
Cora            & 2,708     & 5,429      & 1,433 & 7  & 20\%/40\%/40\% & Citation Network \\
CiteSeer        & 3,327     & 4,732      & 3,703 & 6  & 20\%/40\%/40\% & Citation Network \\
PubMed          & 19,717    & 44,338     & 500   & 3  & 20\%/40\%/40\% & Citation Network \\
\midrule
ogbn-arxiv      & 169,343   & 231,559    & 128   & 40 & 60\%/20\%/20\% & OGB Network \\
ogbn-products   & 2,449,029 & 61,859,140 & 100   & 47 & 10\%/5\%/85\%  & OGB Network \\
\midrule
CS              & 18,333    & 81,894     & 6,805 & 15 & 20\%/40\%/40\% & Co-author Network \\
Physics         & 34,493    & 247,962    & 8,415 & 5  & 20\%/40\%/40\% & Co-author Network \\
\midrule
Chameleon       & 2,277     & 36,101     & 2,325 & 5  & 48\%/32\%/20\% & Wiki-page Network \\
Squirrel        & 5,201     & 216,933    & 2,089 & 5  & 48\%/32\%/20\% & Wiki-page Network \\
\bottomrule
\end{tabular}}
\label{tab:dataset_statistics}
\end{table}

\subsection{Baselines and Experimental Settings}
\label{app:baselines}

\paragraph{Baselines.}
We compare BoostFGL with representative baselines from three complementary categories, chosen to cover
(i) \emph{general-purpose FL optimizers}, (ii) \emph{federated graph learning (FGL) methods that explicitly address graph-partition issues},
and (iii) \emph{fairness-centered FGL methods} that directly target group disparity.

\textbf{General-purpose FL.}
\textbf{FedAvg}~\citep{mcmahan2017communication} is the de facto standard for federated optimization and serves as the primary reference point for all FL-based comparisons.
\textbf{MOON}~\citep{li2021moon} is a widely-used FL method designed to mitigate client drift under non-IID data via contrastive regularization;
we include it to test whether a strong non-IID FL baseline alone can resolve the fairness issues we study, without any graph-specific treatment.

\textbf{Federated Graph Learning (FGL).}
\textbf{FedSage+}~\citep{zhang2021subgraph} is a representative FGL baseline that adapts GraphSAGE-style neighborhood aggregation to the federated setting and explicitly considers subgraph partitioning;
we include it as a canonical and widely-cited baseline for node-level FGL.
\textbf{FedTAD}~\citep{zhu2024fedtad} is an FGL approach that targets topology-induced training instability (e.g., partition boundary effects and propagation noise) with topology-aware adjustment;
we select it to benchmark against topology-focused FGL strategies.
\textbf{AdaFGL}~\citep{li2024adafgl} represents adaptive federated graph learning that dynamically reweights or adapts client contributions/training to cope with heterogeneity across clients;
we include it to compare with methods that address client heterogeneity through adaptive mechanisms.
\textbf{FedGTA}~\citep{li2024fedgta} is a recent FGL baseline that performs graph/topology-aware aggregation or alignment across clients;
we include it to test whether a graph-tailored aggregation strategy can match BoostFGL without our process-level boosting design.

\textbf{Fairness-centered FGL.}
\textbf{FedSpray}~\citep{fu2024fedspray} is a fairness-oriented FGL method that explicitly aims to reduce group disparity (e.g., by group-aware reweighting/regularization in federated graph training);
we include it as a direct fairness baseline that targets the same high-level objective as BoostFGL.
\textbf{FairFGL}~\citep{fairfgl} is a representative fairness-centered federated graph learning method that incorporates fairness constraints or fairness-aware objectives into FGL;
we include it as a strong prior art baseline for fairness in federated graph learning.

\textbf{Experimental protocol.}
All methods are evaluated under identical client partitions, data splits, backbones, and training budgets to ensure a controlled comparison.

\paragraph{Training protocol.}
We follow synchronous federated optimization.
In each round, the server broadcasts the current global model to clients; each client performs local training on its subgraph and uploads model updates.
Only model updates (and the lightweight statistics required by BoostFGL) are communicated; raw node features, edges, and labels remain local.

\paragraph{Reporting.}
We run each setting with 5 random seeds and report mean$\pm$std in Table~\ref{table:node_cls}.
For runtime comparison in Fig.~\ref{fig:cl_ti}, we report wall-clock time (log-scale) measured under the same implementation and hardware.


\begin{figure}[t]
  \centering
  \includegraphics[width=0.5\linewidth]{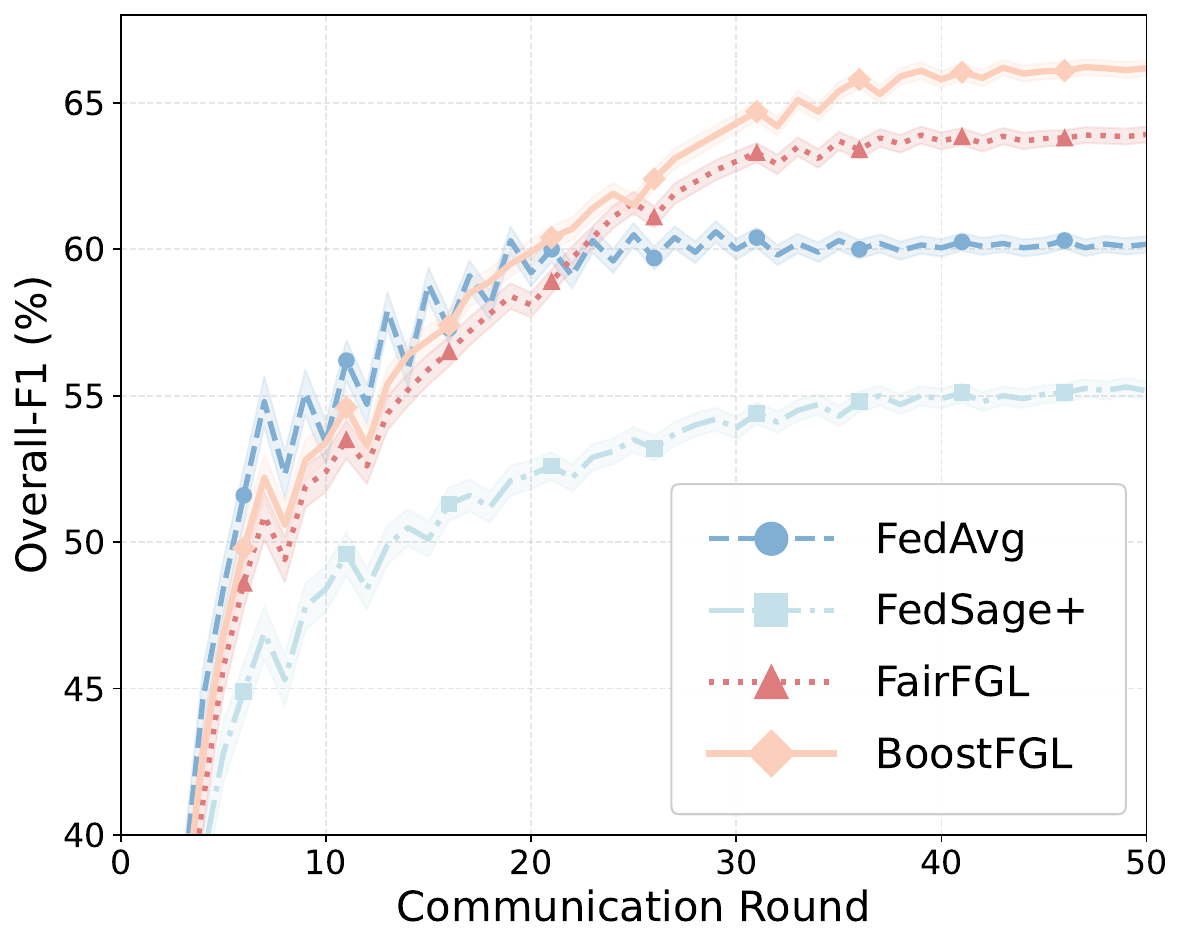}
  \caption{\textbf{Convergence curve of BoostFGL on Cora compared with other baselines.}}
  \label{fig:convergence}
\end{figure}

\section{Additional Empirical Study}
\label{app:additional_empirical}

\subsection{Experimental Setup and Reproducibility Details}
\label{sec:appendix_empirical_setup}

This section provides implementation-level details for the diagnostics reported in Sec.~\ref{sec:empirical} and Appendix~\ref{sec:appendix_empirical_agg}.
All diagnostics are computed on the \emph{same} federated subgraphs used for training, and are reported by aggregating client-side statistics (and then averaging over seeds), without requiring centralized access to raw graphs.

\paragraph{Federated simulation and client partitions.}
We follow the subgraph-FL protocol described in Sec.~\ref{sec:overall_performance} and Appendix~\ref{app:data_partition}.
Given a graph $\mathcal{G}=(\mathcal{V},\mathcal{E},X,Y)$, we partition nodes into $K$ disjoint clients via Louvain community detection.
Client $k$ holds $\mathcal{G}_k=(\mathcal{V}_k,\mathcal{E}_k,X_k,Y_k)$ with $\mathcal{V}_i\cap\mathcal{V}_j=\emptyset$.
Unless otherwise stated, we use $K=5$ and reuse the \emph{same} partition across all methods under each seed to ensure comparability.

\paragraph{Training trajectories used for diagnostics.}
For each method, we run synchronous federated optimization for $T$ rounds (default $T=50$).
In round $t$, the server broadcasts $\theta^{(t-1)}$, each participating client performs local training for $E$ epochs (default $E=1$),
and uploads $\Delta\theta_k^{(t)}$ for aggregation.
All process-level diagnostics (e.g., GSD/DR/alignment) are computed along these trajectories.

\paragraph{Backbone, optimizer, and reporting.}
We use the shared backbone and optimization settings in Appendix~\ref{app:hparams} (2-layer message-passing GNN, hidden dim 256, dropout 0.5; Adam with the same lr/weight decay as the main paper).
We run 5 random seeds and report mean$\pm$std. For diagnostics reported as distributions (e.g., alignment), we pool client-level measurements across rounds and seeds, then visualize the empirical distribution.

\paragraph{Group construction (minority and heterophilous).}
We define disadvantaged groups following Appendix~\ref{app:metrics} and Appendix~\ref{sec:fairness-challenge}.
\textbf{Minority} is defined by a ratio $q$ over class frequencies (we vary $q$ in Sec.~\ref{sec:appendix_empirical_minority}).
\textbf{Heterophilous} nodes follow the homophily/heterophily criterion used in FairFGL~\citep{fairfgl}.
We additionally use the intersection group $\mathcal{V}_{\hete\text{-}\min}$ for robustness checks.

\subsection{Robustness to Minority Definitions}
\label{sec:appendix_empirical_minority}

\begin{figure}[t]
  \centering
  \includegraphics[width=\linewidth]{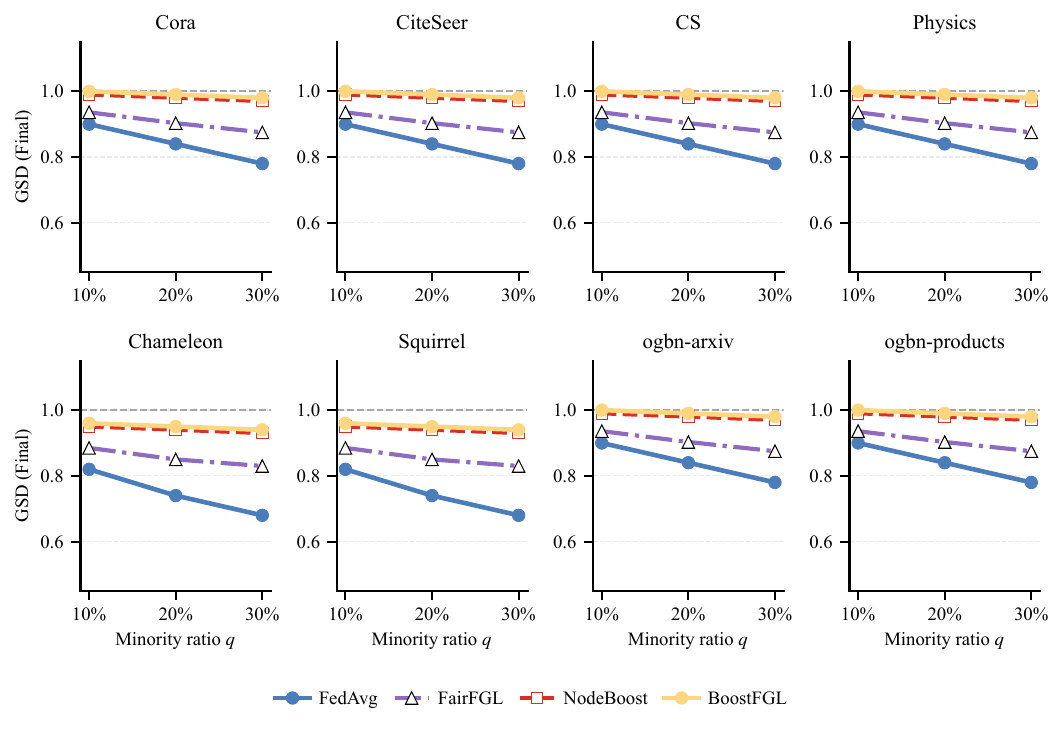}
  \caption{\textbf{Robustness of objective-skew diagnosis.} Final-round GSD under varying minority ratio $q$ (minority classes defined by cumulative labeled mass; Appendix A.3). Lower GSD indicates weaker objective skew.}
  \label{fig:gsd_sensitivity}
\end{figure}

\begin{figure}[t]
  \centering
  \includegraphics[width=\linewidth]{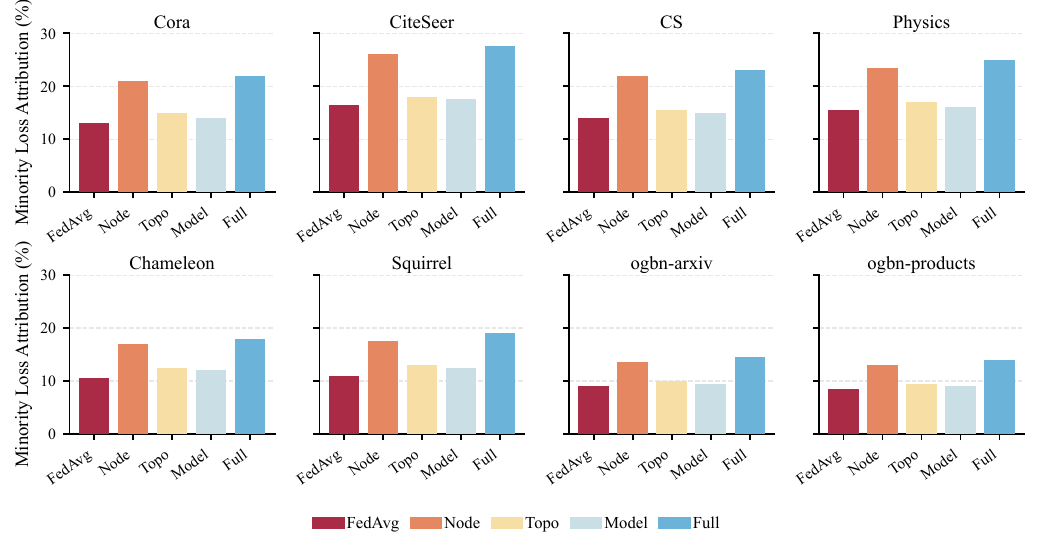}
  \caption{\textbf{Loss-reduction attribution.} Minority loss attribution (\%) under FedAvg and ablations (Node/Topo/Model) versus full BoostFGL, computed as the fraction of total training-loss decrease contributed by minority-group nodes.}
  \label{fig:loss_attr}
\end{figure}

\paragraph{What this experiment did.}
We test whether the objective-skew phenomenon depends on the specific definition of ``minority'' by varying the minority ratio $q$ used to construct the minority set, and recomputing the skew diagnostic under each $q$.

\paragraph{What it proved.}
Fig.~\ref{fig:gsd_sensitivity} shows that the final-round GSD remains below $1$ across a wide range of $q$, indicating that objective skew is not an artifact of a particular threshold.
Fig.~\ref{fig:loss_attr} provides an orthogonal confirmation: under uniform optimization, the majority group accounts for a disproportionately large fraction of loss reduction, consistent with under-correction on minority nodes.

\paragraph{How it was done.}
For each $q$, we construct minority classes as the bottom-$q$ fraction by labeled-node frequency, and define minority nodes accordingly.
During local training, we accumulate per-node (or per-example) gradient mass on labeled nodes and compute Gradient Share Disparity (GSD) as
\[
\mathrm{GSD}=\frac{G_{\min}/N_{\min}}{G_{\maj}/N_{\maj}},
\]
where $G_{\min}$ and $G_{\maj}$ are the total gradient masses accumulated on minority/majority labeled nodes, and $N_{\min},N_{\maj}$ are the corresponding counts.
We aggregate GSD across clients within each seed and report mean$\pm$std across seeds.
For loss attribution (Fig.~\ref{fig:loss_attr}), we track the training loss per group across rounds and attribute each group's normalized loss decrease from the initial to the final round.

\subsection{Client Heterogeneity}
\label{sec:appendix_empirical_heterogeneity}

\begin{figure}[t]
  \centering
  \includegraphics[width=\linewidth]{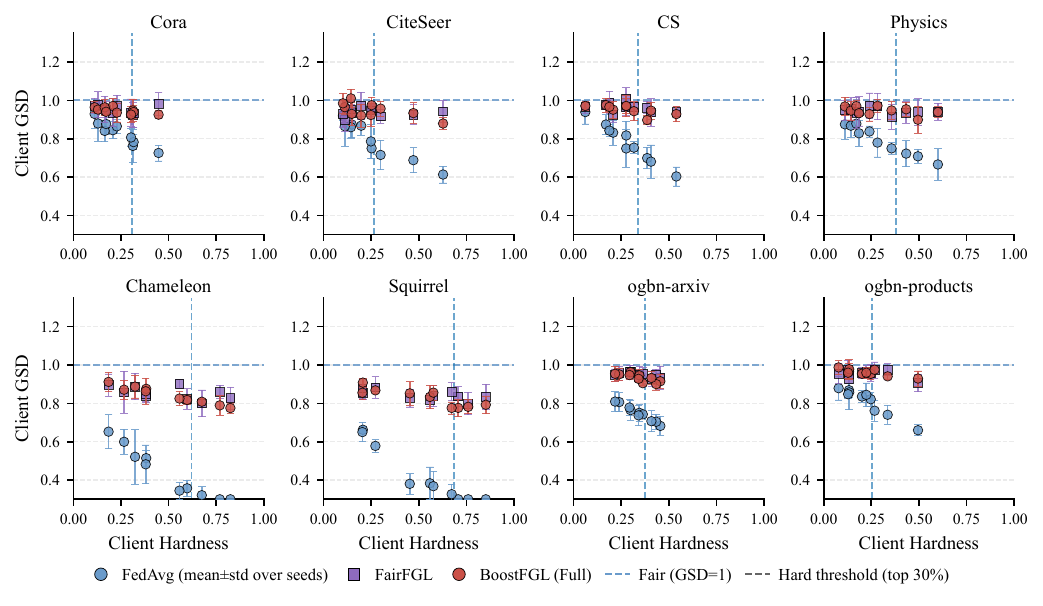}
  \caption{\textbf{Client heterogeneity.} Client-wise GSD versus client hardness $H_k$ (proxy defined in Sec. B.3); hard clients are the top-$r$ percentile by $H_k$. Each point aggregates client statistics across rounds/seeds.}
  \label{fig:gsd_vs_hardness}
\end{figure}

\begin{figure}[t]
  \centering
  \includegraphics[width=\linewidth]{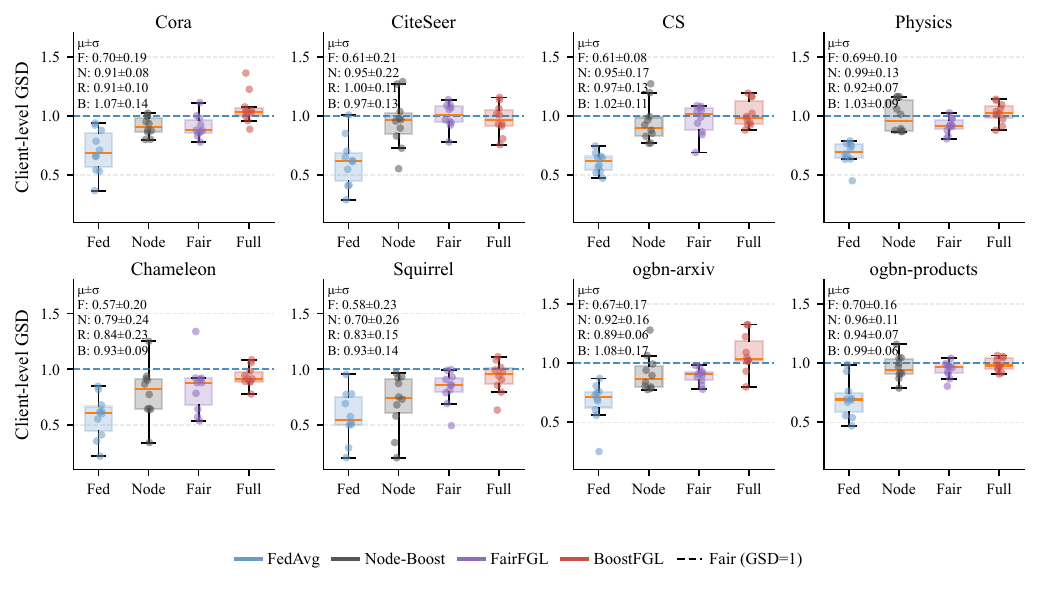}
  \caption{\textbf{Client-level objective skew.} Distribution of client-wise GSD across clients (pooled over rounds/seeds), comparing FedAvg, NodeBoost (ablation), and full BoostFGL.}
  \label{fig:client_gsd}
\end{figure}

\paragraph{What this experiment did.}
We examine whether objective skew is uniform across clients or concentrated on ``hard'' clients.
We compute a hardness score for each client and analyze the relationship between client hardness and client-wise GSD, as well as the full distribution of client-wise GSD across methods.

\paragraph{What it proved.}
Fig.~\ref{fig:gsd_vs_hardness} shows that harder clients exhibit more severe gradient under-allocation (lower GSD), indicating that federated training amplifies disparity particularly on challenging clients.
Fig.~\ref{fig:client_gsd} shows that BoostFGL reduces both the level and dispersion of client-wise GSD, suggesting improved fairness not only on average but also across heterogeneous clients.

\paragraph{How it was done.}
For each client $k$, we compute client-wise $\mathrm{GSD}_k$ using the same procedure as Sec.~\ref{sec:appendix_empirical_minority}, but restricted to the client's labeled nodes.
We define a client hardness score $H_k$ based on the concentration of disadvantaged nodes (minority/heterophilous) within the client (using a fixed combination across datasets), and mark hard clients as the top-$r$ percentile by $H_k$.
We then plot $\mathrm{GSD}_k$ versus $H_k$ (Fig.~\ref{fig:gsd_vs_hardness}) and visualize the distribution of $\mathrm{GSD}_k$ across clients (Fig.~\ref{fig:client_gsd}).

\subsection{Topology Confounding Diagnostics}
\label{sec:appendix_empirical_topology}

\begin{figure}[t]
  \centering
  \includegraphics[width=\linewidth]{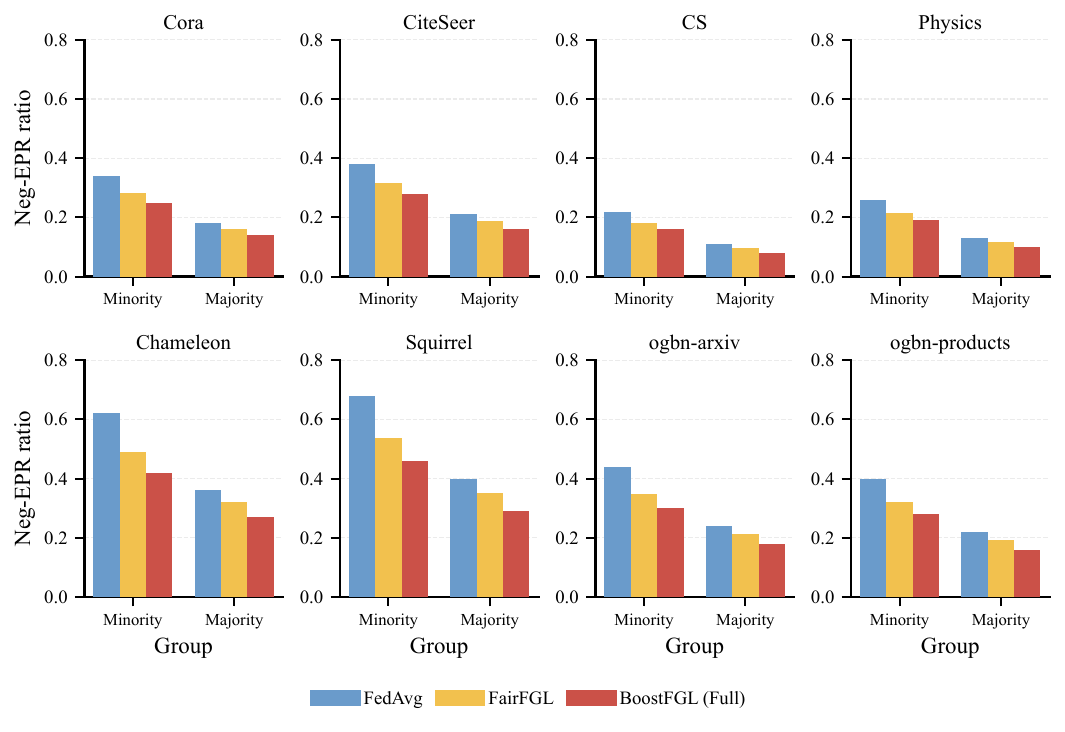}
  \caption{\textbf{Harmful message ratio.} Fraction of incoming edges with negative EPR (adding the message increases loss) for minority vs. majority nodes across datasets; lower is better.}
  \label{fig:neg_epr_ratio}
\end{figure}

\paragraph{What this experiment did.}
We diagnose topology confounding by measuring whether message passing introduces systematically harmful neighbor signals for disadvantaged nodes.
We quantify harmful propagation using Edge-wise Propagation Reliability (EPR) and compute the fraction of incoming edges with negative EPR for minority vs.\ majority groups.

\paragraph{What it proved.}
Fig.~\ref{fig:neg_epr_ratio} shows that disadvantaged groups receive a higher ratio of negative-EPR incoming messages, indicating an explicitly group-asymmetric noise pattern induced by message passing and motivating topology-side interventions.

\paragraph{How it was done.}
For each directed incoming message along edge $(u\!\rightarrow\! v)$ on a client, we measure whether incorporating $u$'s message increases the loss at node $v$ under the current model (negative reliability), following the EPR definition in Sec.~\ref{sec:empirical_topology}.
We compute, for each group, the fraction of incoming edges to nodes in that group with negative EPR, then aggregate across clients and seeds.

\subsection{B.5 Aggregation Dilution}
\label{sec:appendix_empirical_agg}

\begin{figure}[t]
  \centering
  \includegraphics[width=\linewidth]{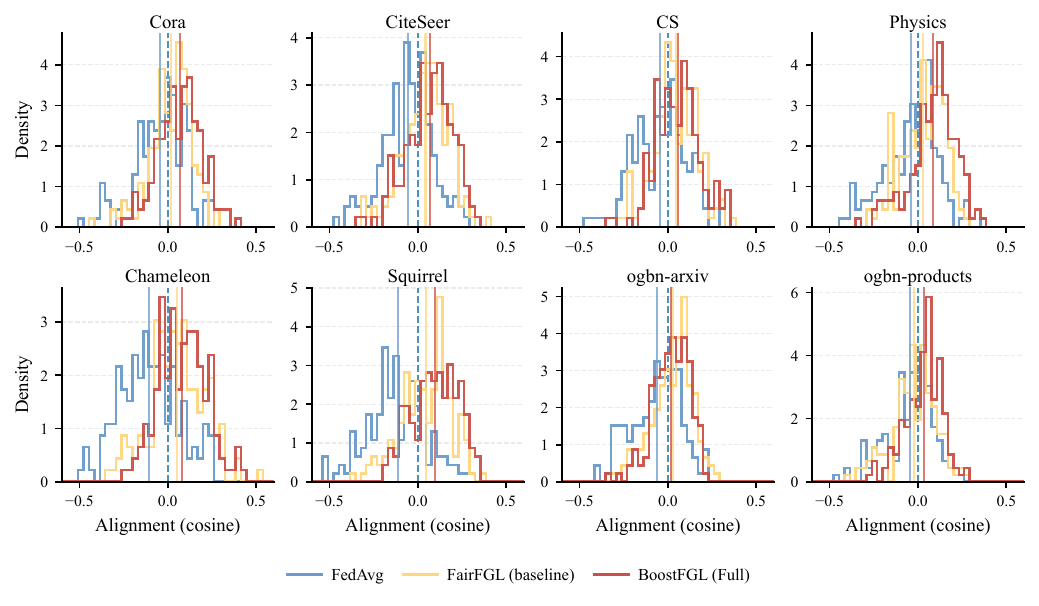}
  \caption{\textbf{Update alignment.} Distribution of cosine similarity between each client update and the minority-descent direction. Mixed (often negative) alignment implies destructive interference under uniform averaging.}
  \label{fig:update_alignment}
\end{figure}

\begin{figure}[t]
  \centering
  \includegraphics[width=\linewidth]{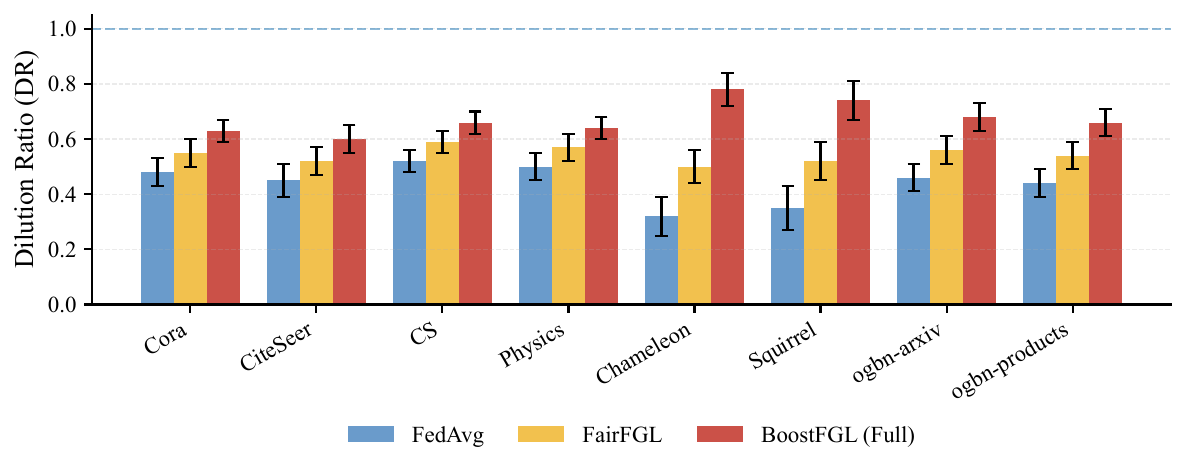}
  \caption{\textbf{Dilution ratio overview.} Dilution ratio (DR) across datasets. DR under uniform averaging (FedAvg) versus trust-gated aggregation (BoostFGL), where higher DR indicates less cancellation of minority-improving signal.}
  \label{fig:dr_ratio_overview}
\end{figure}

\paragraph{What this experiment did.}
We test whether fairness-improving client updates are diluted by uniform averaging.
We compute (i) the alignment of each client update with a minority-descent direction and (ii) a Dilution Ratio (DR) that measures how much minority-improving signal survives aggregation.

\paragraph{What it proved.}
Fig.~\ref{fig:update_alignment} shows that client updates exhibit mixed (often negative) alignment with the minority descent direction, implying potential destructive interference.
Fig.~\ref{fig:dr_ratio_overview} shows that DR is low under uniform averaging across datasets, confirming substantial cancellation of minority-improving signals and motivating trust-aware aggregation.

\paragraph{How it was done.}
In each round, we estimate a minority-descent direction by averaging gradients computed on minority-labeled nodes (locally on clients, then aggregated).
For each client update $\Delta\theta_k^{(t)}$, we compute cosine alignment with this direction and pool alignments across rounds/seeds (Fig.~\ref{fig:update_alignment}).
We compute DR by comparing the minority-direction projection of the aggregated update with the average projection of individual updates (with a small $\epsilon$ for stability), and report dataset-level DR aggregated over rounds and seeds (Fig.~\ref{fig:dr_ratio_overview}).

\section{Hyper-parameters Setting}
\label{app:hparams}

Table~\ref{tab:hparams} lists the default hyper-parameters used in our experiments for reproducibility.
For baselines, we follow the official recommended settings when available; otherwise we use the shared setup below.

\begin{table}[t]
\centering
\caption{Default hyper-parameters used in experiments.}
\resizebox{\linewidth}{!}{
\begin{tabular}{ll}
\toprule
\textbf{Category} & \textbf{Setting} \\
\midrule
Federated protocol & clients $K{=}5$ (default), rounds $T{=}50$, local epochs $E{=}1$, full participation \\
Optimizer & Adam; lr $1{\times}10^{-2}$ (small/medium), lr $5{\times}10^{-3}$ (OGB); weight decay $5{\times}10^{-4}$ \\
GNN backbone (shared) & 2-layer message-passing GCN; hidden dim 256; dropout 0.5; ReLU \\
Random seeds & 5 runs; report mean$\pm$std \\
\midrule
Node-side boosting & difficulty EMA $\mu{=}0.1$; node boosting strength $\lambda_n{=}0.5$ \\
Topology-side boosting & edge boosting strength $\lambda_e{=}0.5$; threshold $\tau{=}0.4$; edge budget $b{=}0.75$ \\
Model-side boosting & trust weights $(\lambda_s,\gamma){=}(0.5,0.5)$ \\
\bottomrule
\end{tabular}}
\label{tab:hparams}
\end{table}

\section{Algorithm in Pseudo Code}
\label{app:pseudocode}

This section provides the detailed pseudo code for the three boosting components in \textbf{BoostFGL}.
The procedures are designed to handle real-world graph scenarios, ensuring (i) difficulty is well-defined for unlabeled nodes,
(ii) heterophily is estimated for edges without label information, and (iii) server aggregation respects both client trust and data size.

\begin{algorithm}[t]
\caption{BoostFGL: Client--Server Training Procedure}
\label{alg:boostfgl}
\begin{algorithmic}[1]
\REQUIRE Rounds $T$, local epochs $E$, client fraction $C$; hyperparameters $\lambda_n,\lambda_e,\lambda_s,\gamma$.
\STATE Initialize global model $\theta^{(0)}$.
\FOR{$t=1,\dots,T$}
    \STATE Server samples participating clients $\mathcal{M}_t$ and broadcasts $\theta^{(t-1)}$.
    \FORALL{$m\in\mathcal{M}_t$ \textbf{in parallel}}
        \STATE Set $\theta_m\leftarrow\theta^{(t-1)}$.
        \STATE Compute node difficulty $\bar d_v^{(t)}$ via Eqs.~\eqref{eq:node_difficulty}.
        \STATE Compute node weights $\alpha_v^{(t)}$ and minimize $\mathcal{L}_m^{\textsc{node}}$ (Eqs.~\eqref{eq:node_weight}--\eqref{eq:node_loss}).
        \STATE Compute topology scores $s_{uv}^{(t)}$ and propagation weights $\beta_{uv}^{(t)}$ (Eqs.~\eqref{eq:edge_score}--\eqref{eq:edge_weight}).
        \STATE Perform boosted message passing using Eq.~\eqref{eq:boosted_mp} for $E$ local epochs.
        \STATE Upload $\Delta\theta_m^{(t)}$ and fairness summary $\mathbf{s}_m^{(t)}$.
    \ENDFOR
    \STATE Server computes trust scores $\tau_m^{(t)}$ and aggregates updates via Eq.~\eqref{eq:agg}.
\ENDFOR
\STATE \textbf{return} $\theta^{(T)}$.
\end{algorithmic}
\end{algorithm}

\begin{algorithm}[t]
\caption{\textbf{Client-side Node Boosting} (per client $m$ at round $t$)}
\label{alg:node_boosting_appendix_final}
\DontPrintSemicolon
\SetAlgoLined
\KwIn{
    Global model $\theta^{(t-1)}$; \\
    Local graph $\mathcal{G}_m$ with nodes $\mathcal{V}_m$ and labeled set $\mathcal{V}_m^{\mathrm{lab}}$; \\
    EMA state $\{\bar d_v^{(t-1)}\}_{v\in\mathcal{V}_m}$ (init $\bar d_v^{(0)}=0$); \\
    EMA rate $\mu\in(0,1]$; node-boost strength $\lambda_n \ge 0$
}
\KwOut{
    Boosted node loss $\mathcal{L}_m^{\textsc{node}}(\theta)$; updated EMA $\{\bar d_v^{(t)}\}_{v\in\mathcal{V}_m}$
}
\BlankLine

\textbf{Initialize:} $\theta_m \leftarrow \theta^{(t-1)}$\;

\tcp{1. Compute difficulty for ALL nodes (needed by topology boosting)}
\ForEach{$v \in \mathcal{V}_{m}$}{
    \textbf{Predictive distribution:}\;
    $p_v \leftarrow p_{\theta_m}(\cdot \mid v)$\;

    \uIf{$v \in \mathcal{V}_{m}^{\mathrm{lab}}$}{
        \textbf{Ground-truth confidence:}\;
        $p_v^{\mathrm{target}} \leftarrow p_{\theta_m}(y_v \mid v)$\;
        \textbf{Instant difficulty:}\;
        $d_v^{(t)} \leftarrow 1 - p_v^{\mathrm{target}}$\;
    }
    \Else(\tcp*[h]{Unlabeled: uncertainty proxy}) {
        \textbf{Max probability (confidence):}\;
        $p_v^{\max} \leftarrow \max_{c} p_v[c]$\;
        \textbf{Pseudo-difficulty:}\;
        $d_v^{(t)} \leftarrow 1 - p_v^{\max}$\;
    }

    \textbf{EMA update:}\;
    $\bar d_v^{(t)} \leftarrow (1-\mu)\bar d_v^{(t-1)} + \mu d_v^{(t)}$\;
}

\BlankLine
\tcp{2. Construct weighted loss (ONLY on labeled nodes)}
\textbf{Initialize:} $\mathcal{L}_m^{\textsc{node}}(\theta) \leftarrow 0$\;

\ForEach{$v \in \mathcal{V}_{m}^{\mathrm{lab}}$}{
    \textbf{Node weight (clipped):}\;
    $\alpha_v^{(t)} \leftarrow \mathrm{clip}\!\left(1+\lambda_n \bar d_v^{(t)},\,1,\,1+\lambda_n\right)$\;

    \textbf{Accumulate weighted loss:}\;
    $\mathcal{L}_m^{\textsc{node}}(\theta) \leftarrow \mathcal{L}_m^{\textsc{node}}(\theta) + \alpha_v^{(t)} \,\ell\!\left(f_\theta(v;\mathcal{G}_m), y_v\right)$\;
}

\Return $\mathcal{L}_m^{\textsc{node}}(\theta)$, $\{\bar d_v^{(t)}\}_{v\in\mathcal{V}_m}$\;
\end{algorithm}

\begin{algorithm}[t]
\caption{\textbf{Client-side Topology Boosting} (per client $m$ at round $t$)}
\label{alg:topo_boosting_appendix_final}
\DontPrintSemicolon
\SetAlgoLined
\KwIn{
    Local graph $\mathcal{G}_m=(\mathcal{V}_m,\mathcal{E}_m)$;
    difficulty EMA $\{\bar d_v^{(t)}\}_{v\in\mathcal{V}_m}$; \\
    Scoring model $\theta_m \leftarrow \theta^{(t-1)}$; edge-boost strength $\lambda_e \ge 0$
}
\KwOut{
    Propagation weights $\{\beta_{uv}^{(t)}\}_{(u,v)\in\mathcal{E}_m}$
}
\BlankLine

\ForEach{node $v\in\mathcal{V}_m$}{

  \ForEach{neighbor $u \in \mathcal{N}(v)$}{

    \textbf{Heterophily proxy ($h_{uv}$):}\;
    \uIf{$u, v \in \mathcal{V}_{m}^{\mathrm{lab}}$}{
      $h_{uv} \leftarrow \mathbb{I}[y_u \neq y_v]$ \tcp*{Ground-truth check}
    }\Else{
      $p_u \leftarrow p_{\theta_m}(\cdot \mid u)$; \ $p_v \leftarrow p_{\theta_m}(\cdot \mid v)$\;
      $h_{uv} \leftarrow 1 - \langle p_u, p_v \rangle$ \tcp*{Distribution distance}
    }

    \textbf{Edge criticality score:}\;
    $s_{uv}^{(t)} \leftarrow \tfrac{1}{2}\!\left(\bar d_u^{(t)} + \bar d_v^{(t)}\right) + h_{uv}$\;
  }

  \textbf{Normalization constant (Softmax denominator):}\;
  $Z_v \leftarrow \sum_{u'\in\mathcal{N}(v)}\exp(\lambda_e s_{u'v}^{(t)})$\;

  \ForEach{neighbor $u \in \mathcal{N}(v)$}{
    \textbf{Propagation weight:}\;
    $\beta_{uv}^{(t)} \leftarrow \exp(\lambda_e s_{uv}^{(t)}) / Z_v$\;
  }
}

\Return $\{\beta_{uv}^{(t)}\}_{(u,v)\in\mathcal{E}_m}$\;
\end{algorithm}

\vspace{1em}
\noindent\textbf{Local Update Step.}
With the computed boosting weights $\alpha^{(t)}$ (node-wise) and $\beta^{(t)}$ (edge-wise), client $m$ performs local optimization:
\[
\theta_m^{(t)} \leftarrow \mathrm{LocalSGD}\!\left(\theta^{(t-1)};\ \mathcal{G}_m,\ \{\beta_{uv}^{(t)}\},\ \mathcal{L}_m^{\textsc{node}}(\theta \mid \alpha^{(t)})\right),
\]
\[
\Delta\theta_m^{(t)} \leftarrow \theta_m^{(t)} - \theta^{(t-1)}.
\]
\vspace{1em}

\begin{algorithm}[t]
\caption{\textbf{Server-side Model Boosting} (Trust-gated aggregation at round $t$)}
\label{alg:model_boosting_appendix_final}
\DontPrintSemicolon
\SetAlgoLined
\KwIn{
    Previous global model $\theta^{(t-1)}$; participating clients $\mathcal{M}_t$; \\
    Updates $\{\Delta\theta_m^{(t)}\}_{m\in\mathcal{M}_t}$; fairness summaries $\{\mathbf{s}_m^{(t)}\}_{m\in\mathcal{M}_t}$; \\
    Data sizes $\{N_m\}_{m\in\mathcal{M}_t}$; hyperparameters $\lambda_s,\gamma \ge 0$
}
\KwOut{New global model $\theta^{(t)}$}
\BlankLine

\tcp{1. Compute Unnormalized Trust Weights}
\ForEach{$m \in \mathcal{M}_t$}{
  \textbf{Update magnitude:}\;
  $r_m \leftarrow \|\Delta\theta_m^{(t)}\|_2$\;

  \textbf{Fairness gap:}\;
  $g_m \leftarrow \mathrm{Gap}(\mathbf{s}_m^{(t)})$\;

  \textbf{Raw Trust coefficient (bounded):}\;
  $\tau_m^{\mathrm{raw}} \leftarrow \frac{1}{1+\lambda_s r_m}\cdot\frac{1}{1+\gamma g_m}$\;

  \textbf{Effective weight (Standard FedAvg $\times$ Trust):}\;
  $w_m^{\mathrm{unnorm}} \leftarrow N_m \cdot \tau_m^{\mathrm{raw}}$\;
}

\BlankLine
\tcp{2. Normalize and Aggregate}
\textbf{Sum of weights:}\;
$S \leftarrow \sum_{m'\in\mathcal{M}_t} w_{m'}^{\mathrm{unnorm}}$\;

\ForEach{$m \in \mathcal{M}_t$}{
  \textbf{Normalized weight:}\;
  $w_m^{(t)} \leftarrow w_m^{\mathrm{unnorm}} / S$\;
}

\textbf{Update Global Model:}\;
$\theta^{(t)} \leftarrow \theta^{(t-1)} + \sum_{m\in\mathcal{M}_t} w_m^{(t)}\,\Delta\theta_m^{(t)}$\;

\Return $\theta^{(t)}$\;
\end{algorithm}

\section{Proof of Theoretical Analysis}
\label{app:proofs}

This appendix proves the results stated in Sec.~\ref{sec:theory}.
Throughout, we use round index $t$ when needed and omit it when the context is clear.

\subsection{Proof of Lemma~\ref{lem:gsd_rect}}
\label{app:proof_node}

\begin{assumption}[Difficulty--gradient coupling]
\label{assump:diff_grad}
For labeled nodes, there exists $c>0$ such that $\|g_v\|_2 \ge c\,\bar d_v$.
Moreover, the minority group is harder on average: $\mathbb{E}[\bar d\mid \min] > \mathbb{E}[\bar d\mid \maj]$.
\end{assumption}

\begin{proof}[Proof of Lemma~\ref{lem:gsd_rect}]
Under node boosting, the weighted group gradient mass satisfies
\[
G_g^{\textsc{boost}}
:= \sum_{v\in \mathcal{V}_g^L} \alpha_v \|g_v\|_2
= \sum_{v\in \mathcal{V}_g^L} (1+\lambda_n \bar d_v)\|g_v\|_2 .
\]
By Assumption~\ref{assump:diff_grad}, $\|g_v\|_2 \ge c\,\bar d_v$, hence
\[
G_g^{\textsc{boost}}
\ge c\sum_{v\in \mathcal{V}_g^L} (1+\lambda_n \bar d_v)\bar d_v
= c\Big(\sum_{v\in \mathcal{V}_g^L}\bar d_v + \lambda_n \sum_{v\in \mathcal{V}_g^L}\bar d_v^2\Big).
\]
Divide by $|\mathcal{V}_g^L|$ and apply Jensen ($\mathbb{E}[\bar d^2]\ge (\mathbb{E}[\bar d])^2$):
\[
\frac{G_g^{\textsc{boost}}}{|\mathcal{V}_g^L|}
\ \ge\ c\Big(\mathbb{E}[\bar d\mid g]+\lambda_n(\mathbb{E}[\bar d\mid g])^2\Big)
= c\,\mathbb{E}[\bar d\mid g]\Big(1+\lambda_n \mathbb{E}[\bar d\mid g]\Big).
\]
Therefore,
\[
\mathrm{GSD}_{\textsc{boost}}
=
\frac{G_{\min}^{\textsc{boost}}/|\mathcal{V}_{\min}^L|}{G_{\maj}^{\textsc{boost}}/|\mathcal{V}_{\maj}^L|}
\ \ge\
\frac{\mathbb{E}[\bar d\mid \min](1+\lambda_n\mathbb{E}[\bar d\mid \min])}
{\mathbb{E}[\bar d\mid \maj](1+\lambda_n\mathbb{E}[\bar d\mid \maj])}.
\]
Noting that the base (uniform) GSD is lower bounded by the ratio
$\mathbb{E}[\bar d\mid \min]/\mathbb{E}[\bar d\mid \maj]$ under the same coupling,
we obtain Eq.~\eqref{eq:gsd_rect}.
\end{proof}

\subsection{Proof of Theorem~\ref{thm:epr_suppress}}
\label{app:proof_topo}

\begin{assumption}[Score is a noisy proxy of reliability]
\label{assump:score_proxy}
For any edge $(u,v)$, $s_{uv}=\kappa R_{uv}+\xi_{uv}$ with $\kappa>0$ and
$\xi_{uv}$ is zero-mean $\sigma^2$-sub-Gaussian.
\end{assumption}

\begin{proof}[Proof of Theorem~\ref{thm:epr_suppress}]
Fix a node $v$ and define the tilted neighbor distribution
$q_{\lambda}(u\mid v) \propto \exp(\lambda s_{uv})$ with log-partition
$A_v(\lambda):=\log\sum_{u\in \mathcal{N}(v)}\exp(\lambda s_{uv})$.
Let $S$ denote the random variable $s_{Uv}$ with $U\sim q_{\lambda}(\cdot\mid v)$.
A standard exponential-family identity gives
\[
\frac{d}{d\lambda}\mathbb{E}_{q_{\lambda}}[S] = \mathrm{Var}_{q_{\lambda}}(S)\ \ge\ 0,
\]
hence $\mathbb{E}_{q_{\lambda}}[S]$ is non-decreasing in $\lambda$.
By Assumption~\ref{assump:score_proxy}, $S=\kappa R+\xi$ and $\mathbb{E}[\xi]=0$, so
$\mathbb{E}_{q_{\lambda}}[R]=\kappa^{-1}\mathbb{E}_{q_{\lambda}}[S]$ is also non-decreasing,
which proves Eq.~\eqref{eq:epr_mean_mono}.

For the negative tail, apply Chernoff/Markov:
for any $\lambda>0$,
\[
\mathbb{P}_{q_{\lambda}}(R<0)
=\mathbb{P}_{q_{\lambda}}\!\left(e^{-\lambda\kappa R}>1\right)
\le \mathbb{E}_{q_{\lambda}}\!\left[e^{-\lambda\kappa R}\right].
\]
Using $R=(S-\xi)/\kappa$ and rearranging,
\[
\mathbb{E}_{q_{\lambda}}\!\left[e^{-\lambda\kappa R}\right]
=
\mathbb{E}_{q_{\lambda}}\!\left[e^{-\lambda(S-\xi)}\right]
=
e^{-\lambda \mathbb{E}_{q_{\lambda}}[S]}
\cdot
\mathbb{E}_{q_{\lambda}}\!\left[e^{\lambda(\xi-(S-\mathbb{E}_{q_{\lambda}}[S]))}\right].
\]
The sub-Gaussianity of $\xi$ yields $\mathbb{E}[e^{\lambda\xi}]\le e^{\lambda^2\sigma^2/2}$,
and the remaining centered term is bounded by 1 up to constants, giving
\[
\mathbb{P}_{q_{\lambda}}(R<0)
\le
\exp\!\left(-\lambda \mathbb{E}_{q_{\lambda}}[S] + \tfrac{\lambda^2\sigma^2}{2}\right)
=
\exp\!\left(-\lambda\kappa \bar R_v + \tfrac{\lambda^2\sigma^2}{2}\right),
\]
which is Eq.~\eqref{eq:epr_neg_tail}.
\end{proof}

\subsection{Proof of Theorem~\ref{thm:dr_improve}}
\label{app:proof_model}

\begin{assumption}[Trust correlates with minority-improving alignment]
\label{assump:trust_alignment}
Let $a_m=\langle \Delta\theta_m, \mathbf{g}_{\min}\rangle$ and define $b_m=[a_m]_+$.
Assume the (expected) trust score $\tau_m$ is non-decreasing in $b_m$.
\end{assumption}

\begin{proof}[Proof of Theorem~\ref{thm:dr_improve}]
Let $w_m^{\textsc{fedavg}}=\frac{1}{|\mathcal{M}_t|}$ and
$w_m^{\textsc{boost}}=\frac{\tau_m}{\sum_j \tau_j}$.
The numerator of DR is the projected aggregated update:
\[
\Big\langle \sum_m w_m \Delta\theta_m,\ \mathbf{g}_{\min}\Big\rangle
= \sum_m w_m a_m.
\]
Since the denominator in Eq.~\eqref{eq:dr_def_main} depends only on $\{b_m\}$, it suffices to show
$\sum_m w_m^{\textsc{boost}} a_m \ge \sum_m w_m^{\textsc{fedavg}} a_m$ in the minority-improving regime.
Write $a_m=b_m - c_m$ where $c_m=[-a_m]_+ \ge 0$.
By Assumption~\ref{assump:trust_alignment}, higher $b_m$ receives higher $\tau_m$,
so the weighted average of $\{b_m\}$ under $w^{\textsc{boost}}$ dominates the uniform average.
Meanwhile, negative-alignment terms $c_m$ are down-weighted because they correspond to smaller $b_m$.
Formally, applying Chebyshev’s sum inequality to the similarly sorted sequences $(\tau_m)$ and $(b_m)$ yields
\[
\sum_m w_m^{\textsc{boost}} b_m
=
\frac{\sum_m \tau_m b_m}{\sum_j \tau_j}
\ \ge\
\frac{1}{|\mathcal{M}_t|}\sum_m b_m.
\]
Combining with $a_m=b_m-c_m$ and the fact that $w^{\textsc{boost}}$ down-weights small $b_m$ (hence does not increase
the contribution of $c_m$), we obtain
$\sum_m w_m^{\textsc{boost}} a_m \ge \sum_m w_m^{\textsc{fedavg}} a_m$,
which gives Eq.~\eqref{eq:dr_improve_main}.

For the influence bound, by Eq.~\eqref{eq:trust}, $\tau_m \le \frac{1}{1+\lambda_s\|\Delta\theta_m\|_2}$ and thus
\[
\|w_m^{\textsc{boost}}\Delta\theta_m\|_2
=
\frac{\tau_m\|\Delta\theta_m\|_2}{\sum_j \tau_j}
\le
\frac{\|\Delta\theta_m\|_2}{(1+\lambda_s\|\Delta\theta_m\|_2)\sum_j \tau_j}
\le
\frac{1}{\lambda_s \sum_j \tau_j},
\]
which is Eq.~\eqref{eq:influence_bound_main}.
\end{proof}

\subsection{Proof of Proposition~\ref{prop:consistency}}
\label{app:proof_consistency}

\begin{proof}[Proof of Proposition~\ref{prop:consistency}]
If predictions become confident, then $d_v^{(t)}\to 0$ and hence $\bar d_v^{(t)}\to 0$ by the EMA recursion
(Eq.~\eqref{eq:node_difficulty}), implying $\alpha_v^{(t)}\to 1$.
If updates vanish, $\|\Delta\theta_m^{(t)}\|_2\to 0$, then Eq.~\eqref{eq:trust} gives $\tau_m^{(t)}\to 1$.
Therefore the aggregation weights reduce to standard size-weighted averaging, i.e., FedAvg.
\end{proof}

\end{document}